\documentclass{article}

\usepackage[preprint,nonatbib]{neurips_2022}

\usepackage[utf8]{inputenc} 
\usepackage[T1]{fontenc}    
\usepackage{hyperref}       
\usepackage{url}            
\usepackage{amsfonts}       
\usepackage{nicefrac}       
\usepackage{microtype}      
\usepackage{xcolor}         
\usepackage{titling}  

\usepackage{booktabs} 
\usepackage[pdftex]{graphicx}
\usepackage{amsfonts}
\usepackage{amsmath}
\definecolor{darkgray1}{gray}{0.33}
\usepackage{colortbl}

\usepackage{amsthm} 
\usepackage{bbm}
\newtheorem{theorem}{Theorem}[section]

\DeclareMathOperator{\E}{\mathbb{E}}
\DeclareMathOperator{\Bin}{Binomial}
\DeclareMathOperator{\Bern}{Bernoulli}
\DeclareMathOperator{\Unif}{Uniform}
\DeclareMathOperator{\Beta}{Beta}
\DeclareMathOperator{\asconv}{\overset{\text{a.s.}}{\to}}

\title{Bridging the Gap: Unifying the Training and Evaluation of Neural Network Binary Classifiers}

\author{
  Nathan Tsoi \\
  Kate Candon\\
  \texttt{nathan.tsoi@yale.edu}
  \and
  Deyuan Li\\
  Yofti Milkessa\\
  Marynel V\'{a}zquez\\
}

\begin{document}

\maketitle

\begin{abstract}
While neural network binary classifiers are often evaluated on metrics such as Accuracy and $F_1$-Score, they are commonly trained with a cross-entropy objective.
How can this training-evaluation gap be addressed? While specific techniques have been adopted to optimize certain confusion matrix based metrics, it is challenging or impossible in some cases to generalize the techniques to other metrics.
Adversarial learning approaches have also been proposed to optimize networks via confusion matrix based metrics, but they tend to be much slower than common training methods.
In this work, we propose a unifying approach to training neural network binary classifiers that combines a differentiable approximation of the Heaviside function with a probabilistic view of the typical confusion matrix values using soft sets.
Our theoretical analysis shows the benefit of using our method to optimize for a given evaluation metric, such as $F_1$-Score, with soft sets, and our extensive experiments show the effectiveness of our approach in several domains.
\end{abstract}

\section{Introduction}

Neural network binary classifiers output a probability $p \in [0,1]$ which is often used at training time to optimize model parameters using the binary cross-entropy (BCE) loss.
The network's output $p$ can also be translated to a binary value $\{0,1\}$ indicating set membership to the negative or positive class.
To determine set membership, the Heaviside step function $H$ is commonly used with a threshold $\tau$, where $p \geq \tau$ are considered positive classification outcomes. This notion of set membership is often used in evaluation metrics for binary classifiers.

It is a common assumption that optimizing a network via the desired evaluation metric is preferable to optimizing a surrogate objective \cite{eban2016scalable, rezatofighi2019generalized, herschtal2004optimising, song2016training}.
However, the Heaviside step function used to compute confusion matrix set membership in terms of true positives, false positives, false negatives, and true negatives, has a gradient with properties not conducive to optimization via gradient descent. The Heaviside function’s gradient is not defined at the threshold $\tau$ and is zero everywhere else.
Therefore the Heaviside function cannot be used to effectively backpropagate errors and train a neural network by optimizing metrics composed of confusion matrix set values, such as $F_1$-Score.

To address the gap between training and evaluating neural network binary classifiers, we propose a method to make confusion-matrix based evaluation metrics usable for backpropagation.
Specifically, we propose the use of a differentiable approximation of the Heaviside step function along with confusion matrix values computed using the notion of soft set membership.
A desired evaluation metric can then be made differentiable by calculating it over the soft-set confusion matrix, rather than using the traditional confusion matrix values.


Our main contributions are: 1) a novel method for training neural network binary classifiers that allows for the optimization of confusion-matrix based evaluation metrics with soft sets (Sec. \ref{sec:method}); 2) a theoretical analysis of our method (Sec. \ref{sec:theory});
and 3) the application of our approach to various domains with varying levels of class imbalance, showing its flexibility and superior performance compared to several baseline methods (Sec. \ref{sec:experiments}). 


\section{Preliminaries}
\label{sec:preliminaries}
In binary classification via neural networks, a step function is required to transform the network's output to a binary value. 
A common choice is the Heaviside step function with a threshold value $\tau$:
{\small
\begin{equation}
H(p, \tau) =
  \begin{cases}
    1 & p \geq \tau \\
    0 & p < \tau
  \end{cases}
\end{equation}
}

Confusion matrix set membership is then computed for a prediction $p$ and ground truth label $y$ via:

{\fontsize{8pt}{0}
\begin{equation}
\begin{alignedat}{2}
\mathit{tp}(p, y, \tau) &= 
  \begin{cases}
    \mathit{H}(p,\tau) & y = 1 \\
    0 & \text{otherwise}
  \end{cases} 
& \qquad \quad \mathit{fn}(p, y, \tau) &= 
  \begin{cases}
    1-\mathit{H}(p,\tau) & y = 1\\
    0 & \text{otherwise}
  \end{cases}\\
\mathit{fp}(p, y, \tau) &= 
  \begin{cases}
    \mathit{H}(p,\tau) & y = 0 \\
    0 & \text{otherwise}
  \end{cases}
&\mathit{tn}(p, y, \tau) &= 
  \begin{cases}
    1-\mathit{H}(p,\tau) & y = 0  \\
    0 & \text{otherwise}
  \end{cases}
\end{alignedat}
\label{eq:set-membership-functions}
\end{equation}
}

Consider a set of predictions $p \in [0,1]$, ground truth labels $y \in \{0,1\}$ and threshold value $\tau \in (0,1)$, e.g., $\{(p_1, y_1, \tau), (p_2, y_2, \tau), ..., (p_n, y_n, \tau)\}$. For $n$ samples, the cardinality of each confusion matrix set is then computed as:
{\small
\begin{equation*}
    |\mathit{TP}| = \sum_{i=1}^n \mathit{tp}(p_i, y_i, \tau)  \quad |\mathit{FN}| = \sum_{i=1}^n \mathit{fn}(p_i, y_i, \tau) \quad
    |\mathit{FP}| = \sum_{i=1}^n \mathit{fp}(p_i, y_i, \tau) \quad |\mathit{TN}| = \sum_{i=1}^n \mathit{tn}(p_i, y_i, \tau)
\end{equation*}
}

Common classification metrics are based on these cardinalities. For example, $\text{Precision}=|\mathit{TP}|/(|\mathit{TP}|+|\mathit{FP}|)$ is the proportion of positive predictions that are true positive results. $\text{Recall}=|\mathit{TP}|/(|\mathit{TP}|+|\mathit{FN}|)$ indicates the proportion of positive examples that are correctly identified. These two metrics represent a trade-off between classifier objectives; it is generally undesirable to optimize or evaluate for one while ignoring the other \cite{he2009learning}. This makes summary evaluation metrics that balance between these trade-offs popular. Commonly used metrics include:
{\small
\begin{align}
\text{Accuracy} = \frac{|\mathit{TP}|+|\mathit{TN}|}{|\mathit{TP}|+|\mathit{TN}|+|\mathit{FP}|+|\mathit{FN}|}
& \qquad \qquad F_1\text{-Score}= \frac{2}{\text{precision}^{-1} + \text{recall}^{-1}}
\end{align}
}

Accuracy is the rate of correct predictions to all predictions.
$F_1$-Score is a specific instance of $F_\beta$-Score, which is the weighted harmonic mean of precision and recall: 
$F_\beta\text{-Score}=(1+\beta^2)\cdot(\text{precision} \cdot \text{recall})/(\beta^2 \cdot \text{precision} + \text{recall})$.
The value $\beta$ indicates that recall is considered $\beta$ times more important than precision.

Some metrics like $F_\beta$-Score are usually computed at a specific threshold $\tau$ whereas others are computed over a range of $\tau$ values. For example, the area under the receiver operating characteristic curve (AUROC) is a commonly used ranking performance measure defined in terms of the true positive rate (TPR) and false positive rate (FPR), each a function of $\tau$: $\int _{\infty }^{-\infty }{\mbox{TPR}}(\tau){\mbox{FPR}}'(\tau) \, d\tau$.

While metrics that rely on confusion matrix set values are commonly used for evaluation, as shown in Fig. \ref{fig:end-to-end} (right), it is difficult to use them as a loss during training.
These metrics rely on the Heaviside step function, the derivative of which is undefined at the threshold $\tau$ and zero everywhere else. This means that these metrics do not have a derivative useful to backpropagate errors through the network. 

When a binary classification neural network is trained on a loss function that is different from the evaluation metric, such as BCE, the network parameters are unlikely to be optimal for the desired evaluation metric. 
As shown in our theoretical analysis and experimental results, when the goal is to balance various confusion matrix set values, e.g., with $F_1$-Score, performance is improved by using our proposed method to optimize $F_1$-Score.

\section{Method}
\label{sec:method}

{\small
\begin{figure*}
    \centering
    \includegraphics[width=\textwidth]{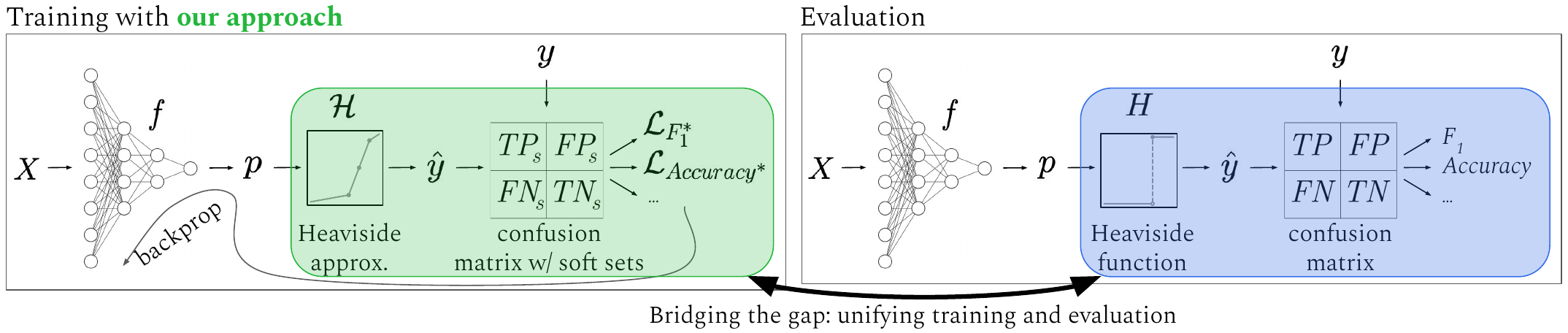}
    \caption{
        The proposed method.
        Binary classifiers are typically trained with the BCE loss and then evaluated on a confusion-matrix based metric.
        We propose to bridge this gap between training and evaluation by optimizing model parameters based on a metric that is computed over a soft-set confusion matrix, which is differentiable.
        This is done using a differentiable approximation of the Heaviside step function.
    }
    \label{fig:end-to-end}
\end{figure*}
}

We propose a method (Fig. \ref{fig:end-to-end}) that aims to better unify the training and evaluation steps of binary neural network classifiers whose performance is measured with metrics based on the confusion matrix.
Our method has two main steps. First, the Heaviside step function, $H$, is approximated with a function $\mathcal{H}$ useful for optimization via gradient descent.
Then, $\mathcal{H}$ is used to compute a soft version of set membership in the confusion matrix.
With the soft-set confusion matrix values, we can then compute desired confusion-matrix based metrics, such as $F_1$-score, using their standard formula.
This appraoch makes the metrics end-to-end differentiable so that they can be used as a training loss.


\subsection{Heaviside Approximation}
\label{sec:heaviside-approximation}

A useful Heaviside approximation, $\mathcal{H}$, has a non-zero gradient: $\mathcal{H}'(p, \tau) \neq 0 \ ,\ \forall \, \tau$.
Also, it ensures that for a single example, the classifier predicts  positive and negative probabilities which sum to 1.
Like $H$, the approximation should meet the properties of a cumulative distribution function (CDF) which ensures it is right-continuous, non-decreasing, with outputs in $[0,1]$ following:
{\small
\begin{equation}
\label{eq:approx-limits}
\lim_{p \to 0} \mathcal{H}(p,\tau) = 0 \hspace{1em} \forall \, \tau \hspace{5em} \lim_{p \to 1} \mathcal{H}(p,\tau) = 1 \hspace{1em} \forall \, \tau
\end{equation}
}

\textbf{Sigmoid approximation:}
One approximation method for $H$, proposed by \cite{kyurkchiev2015sigmoid}, is the sigmoid function $s_0(k;p) = (1+e^{-kp})^{-1}$.
We reparameterize $s_0$ to account for $\tau$, so that it can be used for varying thresholds: $\mathcal{H}^s(k,p,\tau) = (1+e^{-k(p-\tau)})^{-1}$.
A challenge with the sigmoid approximation is that as $k$ increases, the sigmoid function better approximates the Heaviside function, but the derivative is close to zero over a larger range of valid inputs.
Another challenge is that when $\tau$ is close to 0 or 1, $\mathcal{H}^s$ does not approach $H$ as the input approaches 0 or 1.
These limitations as well as the hyperparameters $k$ and $\tau$ are further discussed in the Supplementary Material (Sec. \ref{supp:sec:heaviside-approx}).

\textbf{Linear approximation:}
An alternative approximation of $H$ is a five-point linearly interpolated function (Fig. \ref{fig:end-to-end}, left).
This approximation $\mathcal{H}^l$, is defined over $[0, 1]$ and is parameterized by a given threshold $\tau$ and a slope parameter $\delta$, which define three linear segments with slopes $m_1$, $m_2$, and $m_3$.
The slope of each line segment is:
{\small
    \noindent\begin{minipage}{.33\linewidth}
    \centering
        $$
        m_1 = \frac{\delta}{\tau-\frac{\tau_m}{2}}
        $$
    \end{minipage}%
    \noindent\begin{minipage}{.33\linewidth}
    \centering
        $$
        m_2 = \frac{1-2\delta}{\tau_m}
        $$
    \end{minipage}%
    \noindent\begin{minipage}{.33\linewidth}
    \centering
        $$
        m_3 = \frac{\delta}{1-\tau-\frac{\tau_m}{2}}
        $$
    \end{minipage}
}
with $\tau_m = \text{min} \{\tau, 1-\tau\}$ in order to ensure a gradient suitable for backpropagation.
The linear Heaviside approximation is thus given by:
{\small
\begin{equation}
\label{eq:linear}
\mathcal{H}^l =
  \begin{cases}
    p \cdot m_1 & \text{if $p < \tau - \frac{\tau_m}{2}$} \\
    p \cdot m_3 +(1-\delta-m_3(\tau+\frac{\tau_m}{2})) & \text{if $p > \tau + \frac{\tau_m}{2}$} \\
    p \cdot m_2 + (0.5 - m_2\tau) & \text{otherwise}
  \end{cases}
\end{equation}
}

Considering the threshold $\tau$ in the formulation of $\mathcal{H}^l$ ensures $\mathcal{H}^l(p=\tau, \tau)=0.5$ while maintaining the limits in Eq. \eqref{eq:approx-limits}.
See the Supplementary Material, Sec. \ref{supp:sec:linear-derivation}, for the derivation of Eq. \eqref{eq:linear}.

Recently, a similar linear function was implemented as an activation function to enforce meaningful logical outputs in logical neural networks \cite{riegel2020logical}. However, rather than using the linear function for activation within the larger model architecture, we use it in the training objective of a classifier.

\subsection{Soft Sets}
We use soft sets, a generalization of fuzzy sets \cite{molodtsov1999soft}, to compose differentiable versions of confusion-matrix based evaluation metrics useful for backpropagation.
A soft set $\mu$ in $U$, the initial universal set, is defined by the membership function $\mu: U \to [0,1]$. For $x \in U$, the set membership function $\mu(x)$ specifies a degree of belonging for $x$ to set $u$. 

We use the notion of soft sets to compute confusion-matrix based metrics with a Heaviside approximation in place of the typically used strict sets.
Soft set membership corresponds to the degree to which a sample tuple $(p, y, \tau)$ belongs to a confusion matrix set.
We define the soft confusion matrix set membership functions for prediction and ground truth examples $(p, y)$ relative to $\tau$ as:

{\fontsize{8pt}{0}
\begin{equation}
\begin{alignedat}{2}
\mathit{tp_s}(p, y, \tau) &= 
  \begin{cases}
    \mathcal{H}(p,\tau) & y = 1 \\
    0 & \text{otherwise}
  \end{cases} 
& \qquad \quad \mathit{fn_s}(p, y, \tau) &= 
  \begin{cases}
    1-\mathcal{H}(p,\tau) & y = 1\\
    0 & \text{otherwise}
  \end{cases}\\
\mathit{fp_s}(p, y, \tau) &= 
  \begin{cases}
    \mathcal{H}(p,\tau) & y = 0 \\
    0 & \text{otherwise}
  \end{cases}
&\mathit{tn_s}(p, y, \tau) &= 
  \begin{cases}
    1-\mathcal{H}(p,\tau) & y = 0  \\
    0 & \text{otherwise}
  \end{cases}
\end{alignedat}
\label{eq:soft-set-membership-functions}
\end{equation}
}

After computing per-sample soft set values, specific metrics can be approximated by summing over the relevant elements of the confusion matrix.
For instance, we approximate precision as $|\mathit{TP_s}|/(|\mathit{TP_s}|+|\mathit{FP_s}|)$, where  $|\mathit{TP_s}| = \sum_{i=1}^n \mathit{tp_s}(p_i, y_i, \tau)$ and $|\mathit{FP_s}| = \sum_{i=1}^n \mathit{fp_s}(p_i, y_i, \tau)$ for $n$ samples. In practice, this summation occurs at training time over mini-batches while optimizing via gradient descent. Because gradient descent and its variants expect a small but representative sample of the broader data \cite{bottou2012stochastic}, the proposed method also expects a representative sample.

\textbf{Decomposability:} 
Metrics based on confusion matrix set values are considered non-decomposable. Non-decomposable metrics cannot be calculated per datapoint and are not additive across subsets of data \cite{kar2014online}.
We acknowledge that our optimization method of mini-batch stochastic gradient descent (SGD) does not provide an unbiased estimator on non-decomposable metrics.
However, this is common in neural network training \cite{goodfellow2016deep}.
In practice, we find that large enough mini-batches provide a representative sample for confusion matrix based metrics, allowing the use of mini-batch SGD in our approach.
Moreover, our method does not limit maximum batch size for training, unlike the adversarial approach to optimize $F_1$-Score proposed by \cite{fathony2020ap}.

\textbf{Metrics and losses:} Any metric composed of confusion matrix set values ($\mathit{TP}$, $\mathit{FP}$, $\mathit{FN}$, $\mathit{TN}$) can be approximated using our proposed method and used as a training loss.
In our experiments, we train and evaluate on Accuracy, $F_1$-Score, AUROC, and $F_\beta$-Score  \cite{Rijsbergen}. 
We chose these metrics to show the flexibility of our approach. Accuracy, $F_1$-Score and $F_\beta$-Score may be selected based on class imbalance and resulting tradeoffs. AUROC illustrates that our method can be used when computation over a range of thresholds, $\tau$, is required.

\section{Theoretical Grounding}
\label{sec:theory}
In this section, we first show the Lipschitz continuity of a variety of soft-set based metrics under the proposed Heaviside function approximations. This indicates that when such metrics are used as the loss in a neural network, the difference between successive losses is bounded across iterations of stochastic gradient descent.
We then show, under certain assumptions, that metrics computed over the soft-set confusion matrix values are asymptotically similar to the true metric.

\subsection{Lipschitz Continuity of Metrics Based on Soft-Set Confusion Matrix Values}
\label{sec:theory:continuity}

A function $f$ is considered Lipschitz continuous if there exists some constant $K$ such that for all $x_1, x_2$ in the domain, $|f(x_1)-f(x_2)| \le K|x_1 - x_2|$. Lipschitz continuous functions are always themselves continuous. 

\begin{theorem} \label{main:thm:HLip}
The linear Heaviside approximation $\mathcal{H}^l$ is Lipschitz continuous with Lipschitz constant $M = \max \{m_1, m_2, m_3\}$. 
\end{theorem}

$\mathcal{H}^l$ is continuous because each piecewise linear component is continuous and $\mathcal{H}^l$ is defined to be continuous at the two points $p = \tau \pm \frac{\tau_m}{2}$. Also, by construction, the slope of any secant line is positive and bounded by $M = \max \{m_1, m_2, m_3 \}$. Thus, $\mathcal{H}^l$ is Lipschitz continuous with Lipschitz constant $M$.
Please see the full proof in Sec. \ref{supp:sec:theory:lipschitz} of the Supplementary Material.
Note that the $\mathcal{H}^s$ 
is also Lipschitz continuous \cite{shang2021lipschitz}. 


\begin{theorem} \label{main:thm:SoftLip}
Every entry of the soft-sets confusion matrix based on the Heaviside approximations is Lipschitz continuous in the output of a neural network.
\end{theorem}

The proof of Theorem \ref{main:thm:SoftLip} is in the Supplementary Material (Sec. \ref{supp:sec:theory:lipschitz}). 

The value of a Lipschitz continuous loss function composed of soft-set confusion matrix values is Lipschitz continuous. This is because compositions of Lipschitz continuous functions are also Lipschitz continuous.
Following \cite{dembczynski2017consistency}, we note that the confusion-matrix based metrics Accuracy, Balanced Accuracy, $F_\beta$, Jaccard, and G-Mean are all Lipschitz continuous.
These metrics are also Lipschitz continuous under our proposed method computed via soft sets. 
%

%
%

Lipschitz continuity of the loss function in a neural network optimized via stochastic gradient descent indicates convergence without extreme variations in losses throughout training.
Let $K$ be the Lipschitz constant for the objective function $\ell(w)$ on network weights $w$. Because of Lipschitz continuity, 
when updating $w_{i+1} \to w_i - \alpha_i \ell'(w_i)$ using stochastic gradient descent with a learning rate $\alpha_i$, a small local change in the weights $|w_{i+1} - w_i| = \alpha_i |\ell'(w_i)|$ corresponds to a small local change in the value of the objective function of $|\ell(w_{i+1}) - \ell(w_i)| \le K\alpha_i|\ell'(w_i)|$. 
\subsection{Approximation of Confusion-Matrix Based Metrics with Soft Sets}
\label{sec:theory:convergence}

We provide a statistical analysis showing that in the limit, as the number of examples goes to infinity, $F_1$-Score calculated with soft sets approximates the expected true $F_1$-score under a set of assumptions. Similar proofs for Accuracy and AUROC are provided in the Supplementary Material (Sec. \ref{supp:sec:theory:converge}). 


Consider a dataset of size $n$ with $\{x_1, ..., x_n\}$ examples and $\{y_1, ..., y_n\}$ labels.
Suppose a network outputs a probability $p_i$ that the label $y_i=1$. In this section, since the specific outputs $p_i$ are unknown and may change across iterations, we assume $p_i$ is a random variable.
When calculating $F_1$-Score, $p_i$ is passed through the Heaviside function $H$, which generates an output $\hat y_i^H = H(p_i, \tau)$, where $\hat y_i^H \in \{0, 1\}$. 
The $F_1$-Score can be expressed as:
{\small
\begin{equation}
\label{eq:f1-yhat}
\begin{alignedat}{2}
    F_1 
    &= \frac{\sum_{i=1}^n y_i\hat y_i^H}{\sum_{i=1}^n y_i\hat y_i^H + \frac{1}{2}\sum_{i=1}^n ((1-y_i)\hat y_i^H + y_i(1-\hat y_i^H))}
    &=\frac{2\sum_{i=1}^n y_i\hat y_i^H}{\sum_{i=1}^n (y_i + \hat y_i^H)}
\end{alignedat}
\end{equation}
}

To calculate $F_1$-Score with soft sets ($F_1^s$), $p_i$ is passed through the Heaviside approximation $\mathcal{H}$, which generates an output $\hat y_i^{\mathcal{H}} = \mathcal{H}(p_i, \tau)$, where $\hat y_i^{\mathcal{H}} \in [0, 1]$.
$F_1$-Score with soft-sets can be expressed as: 
{\small
\begin{equation}
\label{eq:f1s-yhat}
\begin{alignedat}{2}
    F_1^s 
    &=\frac{2\sum_{i=1}^n y_i\hat y_i^{\mathcal{H}}}{\sum_{i=1}^n (y_i + \hat y_i^{\mathcal{H}})}
\end{alignedat}
\end{equation}
}

Consider a network trained on a dataset with $rn$ positive and $(1-r)n$ negative elements, where $r \in [0,1]$ is some constant. Suppose this classifier correctly classifies any positive example as a true positive with probability $u$ and any negative example as a false positive with probability $v$.
Also, assume that all classifications are independent.
Because $F_1$-Score is calculated with discrete $\hat y_i^H$, we assume that the classifier will classify examples as a random variable $\hat y_i^H \sim \Bern(uy_i + v(1-y_i))$. 
Thus, $\hat y_i^H \sim \Bern(u)$ if $y_i=1$, and $\hat y_i^H \sim \Bern(v)$ if $y_i=0$. 

Since $F_1^s$ can take on continuous values in $[0, 1]$, we consider that $\hat y_i^{\mathcal{H}}$ is a random variable drawn from a Beta distribution, which has support $[0, 1]$. In particular, assume $\hat y_i^{\mathcal{H}} \sim \Beta(\alpha_uy_i + \alpha_v(1-y_i), \beta_uy_i + \beta_v(1-y_i))$. Hence, $\hat y_i^{\mathcal{H}} \sim \Beta(\alpha_u, \beta_u)$ if $y_i=1$, and $\hat y_i^{\mathcal{H}} \sim \Beta(\alpha_v, \beta_v)$ if $y_i=0$. Let $\frac{\alpha_u}{\alpha_u + \beta_u} = u$ and $\frac{\alpha_v}{\alpha_v+\beta_v} = v$, so for any $i$, $\E\left[\hat y_i^{\mathcal{H}}\right]=u$ if $y_i = 1$, and $\E\left[\hat y_i^{\mathcal{H}}\right]=v$ if $y_i = 0$. 

Under the above assumptions, both $F_1$ and $F_1^s$ have the same average classification correctness: for any given $i$, $\E\left[\hat y_i^H\right] = \E\left[\hat y_i^{\mathcal{H}}\right]$. Also, there exists some $\alpha_u, \beta_u, \alpha_v, \beta_v$ such that the distributions of $\hat y_i^H = H(p_i, \tau)$ and $\hat y_i^{\mathcal{H}} = \mathcal{H}(p_i, \tau)$ can both hold simultaneously under the same network for all $i$.

If we let $U \sim \Bin(nr, u)$ and $V \sim \Bin(n(1-r), v)$ be the independent random variables denoting the number of true positives and false positives in a sequence of $n$ independent predictions, then $F_1$ from Eq. \eqref{eq:f1-yhat} becomes:
{\small
\begin{equation}
\label{eq:f1-U}
\begin{alignedat}{2}
F_1 =\frac{2\sum_{i=1}^n y_i\hat y_i^H}{\sum_{i=1}^n (y_i + \hat y_i^H)} = 2\left(\frac{\sum_{i=1}^n y_i\hat y_i^H}{nr + \sum_{i=1}^n \hat y_i^H}\right) = 2\left(\frac{U}{nr + U+V}\right) = 2\left(\frac{U/n}{r + U/n + V/n}\right)
\end{alignedat}
\end{equation}
}

By the Strong Law of Large Numbers, 
$\frac{1}{nr}U \asconv u$ and $\frac{1}{n(1-r)}V \asconv v$ both converge with probability $1$ as $n \to \infty$.
Hence, $\frac{U}{n} \asconv ru$ and $\frac{V}{n} \asconv (1-r)v$.
We therefore have, from the Continuous Mapping Theorem, that as $n \to \infty$:

{\small
\begin{equation}
\label{eq:f1-littleu}
\begin{alignedat}{2}
F_1 = 2\left(\frac{U/n}{r + U/n + V/n}\right) \asconv 2\left(\frac{ru}{r + ru + (1-r)v}\right) = \frac{2ru}{r + ru + v - rv}
\end{alignedat}
\end{equation}
}



For $F_1^s$, let $U^s=\sum_{y_i = 1} \hat y_i^{\mathcal{H}}$ and $V^s = \sum_{y_i = 0} \hat y_i^{\mathcal{H}}$ be the independent random variables denoting the total amount of true positives and false positives in the soft set case. Then, $F_1^s$ from Eq. \eqref{eq:f1s-yhat} becomes:
{\small
\begin{equation}
\label{eq:f1s-U}
\begin{alignedat}{2}
F_1^s =\frac{2\sum_{i=1}^n y_i\hat y_i^{\mathcal{H}}}{\sum_{i=1}^n (y_i + \hat y_i^{\mathcal{H}})}
= 2\left(\frac{\sum_{i=1}^n y_i\hat y_i^{\mathcal{H}}}{nr + \sum_{i=1}^n \hat y_i^{\mathcal{H}}}\right) = 2\left(\frac{U^s}{nr + U^s+V^s}\right) = 2\left(\frac{U^s/n}{r + U^s/n + V^s/n}\right)
\end{alignedat}
\end{equation}
}

Since $U^s$ is the sum of $nr$ i.i.d. random variables distributed as $\Beta(\alpha_u, \beta_u)$, which has mean $\frac{\alpha_u}{\alpha_u + \beta_u}$, by the Strong Law of Large Numbers, 
$\frac{1}{nr}U^s \asconv \frac{\alpha_u}{\alpha_u + \beta_u} = u$. Similarly, $\frac{1}{n(1-r)}V^s \asconv \frac{\alpha_v}{\alpha_v+\beta_v}=v$ also converges with probability $1$ as $n \to \infty$.
Hence, $\frac{U^s}{n} \asconv ru$ and $\frac{V^s}{n} \asconv (1-r)v$.
We therefore have, from the Continuous Mapping Theorem, that as $n \to \infty$,

{\small
\begin{equation}
\label{eq:f1s-littleu}
\begin{alignedat}{2}
F_1^s = 2\left(\frac{U^s/n}{r + U^s/n + V^s/n}\right) \asconv 2\left(\frac{ru}{r + ru + (1-r)v}\right) = \frac{2ru}{r + ru + v - rv}
\end{alignedat}
\end{equation}
}


Thus, $F_1$ and $F_1^s$ both converge almost surely to the same value as $n \to \infty$. 
Since $0 \le \frac{2ru}{r + ru + v - rv} \le 1$ is bounded, $\E[F_1], \E[F_1^s] \to \frac{2ru}{r + ru + v - rv}$ by the Bounded Convergence Theorem. This means that the $F_1^s$ value is an asymptotically unbiased estimator for the expected true $F_1$-Score, and we expect average $F_1$-Score values to converge to $F_1^s$ as $n \to \infty$, under our setup.
However, the $F_1$-Score computed from soft sets is not an unbiased estimator for the expected true $F_1$-Score for finite $n$. 

The proof of almost sure convergence generalizes for any metric that is a continuous function in the ratio of each entry of the confusion matrix to $n$,
as shown in the Supplementary Material (Sec. \ref{supp:sec:theory:converge}). 
This suggests that optimizing over the desired evaluation metric using our proposed method (e.g., $F_1$-Score computed with soft sets) is justified, as the loss should follow our true final loss closely for a large enough $n$.
However, note that some metrics may be poor indicators of true classifier performance, such as Accuracy when data is imbalanced \cite{he2009learning}.
In such cases, it may not be desirable to optimize for such metrics using our method, as further discussed in the next section.

\section{Experiments}
\label{sec:experiments}
This section presents three experiments to 
(1) evaluate the performance of our approach against several baselines on tabular data, 
(2) evaluate our approach on higher-dimensional image data,
and
(3) evaluate the ability of our method to balance precision and recall during training.

\textbf{Datasets:}
Experiments were conducted on five publicly available datasets in a variety of domains and were chosen for their varying levels of class imbalance as explained later.
All datasets were minimally pre-processed and split into separate train, test, and validation sets.
See Sec. \ref{supp:sec:data} of the Supplementary Material for details.

\textbf{Architecture and training:}
Our experiments aim to fairly evaluate our method using different confusion-matrix based objective functions.
Therefore, the same network architecture and training scheme was used unless otherwise noted.
Performance was evaluated over 10 repeated trials to control for the effects of random weight initialization.
We report the mean of results evaluated over threshold values $T = \{0.1, 0.2, ..., 0.9\}$ for metrics that require a threshold choice at evaluation. 
The details of the network architecture and training scheme are in Sec. \ref{supp:sec:arch-train} of the Supplementary Material.

\textbf{Baselines:}
We trained networks using the typical BCE loss as well as two existing approaches for optimizing specific confusion matrix metrics: an adversarial approach for $F_1$-score \cite{fathony2020ap}, and using an approximation of the Wilcoxon-Mann-Whitney (WMW) statistic for AUROC \cite{yan2003optimizing}.
Also, the Supplementary Material provides comparisons with other, less related methods for binary classification.

\subsection{Experiments on Tabular Data}
\label{sec:direct-optimization-experiment}

\begin{table*}[tb!]
\setlength{\tabcolsep}{4pt}
\caption{
  Losses (rows): $F_1$, Accuracy ($Acc$), and AUROC ($ROC$) via the proposed method (*) using the linear approximation;
  $F_1$-Score\dag\ via adversarial approach \cite{fathony2020ap} and AUROC\ddag\ via WMW statistic \cite{yan2003optimizing}. Bold indicates performance better than or equal to the BCE baseline.
}
\begin{center}
{\small
\begin{tabular}{lrcccccc}
\toprule
\multicolumn{2}{c}{} & \multicolumn{3}{c}{\textbf{CocktailParty} ($\mu\pm\sigma$)} & \multicolumn{3}{c}{\textbf{Adult} ($\mu\pm\sigma$)} \\ \cmidrule(lr){3-5} \cmidrule(lr){6-8}
& Loss & $F_1$-Score & Accuracy & AUROC&$F_1$-Score & Accuracy & AUROC \\ \midrule
(1) & $\text{F}_1$* & $\mathbf{0.75 \pm 0.01}$ & $\mathbf{0.85 \pm 0.01}$ & $\mathbf{0.82 \pm 0.01}$ & $\mathbf{0.63 \pm 0.02}$ & $0.78 \pm 0.04$ & $\mathbf{0.78 \pm 0.02}$ \\
(2) & $\text{F}_1$\dag & $0.30 \pm 0.06$ & $0.76 \pm 0.01$ & $0.60 \pm 0.02$ & $0.16 \pm 0.02$ & $0.78 \pm 0.00$ & $0.55 \pm 0.01$ \\
(3) & $\text{Accuracy}$* & $\mathbf{0.70 \pm 0.02}$ & $\mathbf{0.85 \pm 0.01}$ & $\mathbf{0.78 \pm 0.01}$ & $\mathbf{0.35 \pm 0.04}$ & $\mathbf{0.81 \pm 0.01}$ & $\mathbf{0.61 \pm 0.02}$ \\
(4) & $\text{AUROC}$* & $0.51 \pm 0.01$ & $0.41 \pm 0.01$ & $0.57 \pm 0.00$ & $\mathbf{0.42 \pm 0.01}$ & $0.32 \pm 0.02$ & $0.55 \pm 0.01$ \\
(5) & AUROC\ddag & $0.01 \pm 0.03$ & $0.70 \pm 0.03$ & ${\color{black}0.50 \pm 0.00}$ & $0.00 \pm 0.00$ & $0.76 \pm 0.00$ & ${\color{black}0.50 \pm 0.00}$ \\
(6) & BCE & $0.70 \pm 0.02$ & $0.85 \pm 0.01$ & $0.78 \pm 0.01$ & $0.26 \pm 0.06$ & $0.80 \pm 0.01$ & $0.58 \pm 0.02$ \\
\end{tabular}
}
{\small
\begin{tabular}{lrcccccc}
\toprule
\multicolumn{2}{c}{} & \multicolumn{3}{c}{\textbf{Mammography} ($\mu\pm\sigma$)} & \multicolumn{3}{c}{\textbf{Kaggle} ($\mu\pm\sigma$)} \\ \cmidrule(lr){3-5} \cmidrule(lr){6-8}
& Loss & $F_1$-Score & Accuracy & AUROC&$F_1$-Score & Accuracy & AUROC \\ \midrule
(1) & $\text{F}_1$* & $\mathbf{0.63 \pm 0.04}$ & $0.98 \pm 0.00$ & $\mathbf{0.78 \pm 0.03}$ & $\mathbf{0.83 \pm 0.02}$ & $\mathbf{1.00 \pm 0.00}$ & $\mathbf{0.90 \pm 0.02}$ \\
(2) & $\text{F}_1$\dag & $0.46 \pm 0.08$ & $0.98 \pm 0.00$ & $0.66 \pm 0.04$ & $\mathbf{0.76 \pm 0.06}$ & $\mathbf{1.00 \pm 0.00}$ & $\mathbf{0.83 \pm 0.04}$ \\
(3) & $\text{Accuracy}$* & $0.00 \pm 0.00$ & $0.97 \pm 0.00$ & $0.50 \pm 0.00$ & $\mathbf{0.62 \pm 0.33}$ & $\mathbf{1.00 \pm 0.00}$ & $\mathbf{0.78 \pm 0.15}$ \\
(4) & $\text{AUROC}$* & $0.11 \pm 0.01$ & $0.18 \pm 0.04$ & $0.57 \pm 0.02$ & $0.06 \pm 0.01$ & $0.11 \pm 0.00$ & $0.55 \pm 0.00$ \\
(5) & AUROC\ddag & $0.00 \pm 0.01$ & $0.88 \pm 0.12$ & ${\color{black}0.50 \pm 0.00}$ & $0.00 \pm 0.00$ & $0.93 \pm 0.15$ & ${\color{black}0.50 \pm 0.00}$ \\
(6) & BCE & $0.56 \pm 0.11$ & $0.99 \pm 0.00$ & $0.71 \pm 0.06$ & $0.50 \pm 0.33$ & $1.00 \pm 0.00$ & $0.73 \pm 0.16$ \\
\end{tabular}
}
\end{center}
\label{tbl:direct-results}
\vspace{-2em}
\end{table*}

We evaluate our proposed approach using four tabular datasets with different levels of class imbalance. 
The CocktailParty dataset has a 30.29\% positive class balance making it the most class-balanced dataset of those considered in this experiment. 
Salary classification data in the Adult dataset has a 23.93\% positive class balance.
Classifications of microcalcifications in the Mammography dataset is heavily skewed with a 2.32\% positive class balance.
Lastly, the Kaggle Credit Card Fraud Detection dataset has the most extreme class balance with only a 0.17\% positive class balance.

We compare baseline methods against the performance of neural networks trained using our method to optimize $F_{1}$-Score, Accuracy, and AUROC. 
We instantiated our method using the Heaviside approximations and soft sets.
Results using the linear approximation are presented in Table \ref{tbl:direct-results}.
See the Supplementary Material, Sec. \ref{supp:sec:direct-optimization-experiment}, for results using the sigmoid approximation, which were comparable.


\textit{$F_1$ computed with soft sets.}
Our method of optimizing $F_1$ over the soft-set confusion matrix outperforms baselines when evaluated on $F_1$-Score for all datasets. In Table \ref{tbl:direct-results}, line (1) has a higher $F_1$-Score than lines (2) and (6).
Additionally, even when evaluated on the other metrics (Accuracy and AUROC), networks trained using our method on $F_1$ perform similarly or better than BCE.

\textit{Accuracy computed with soft sets.}
Our method of optimizing Accuracy over the soft-set confusion matrix has comparable performance to the BCE baseline when evaluated on Accuracy for all datasets.
In Table \ref{tbl:direct-results}, line (3) has comparable Accuracy to line (6).
Note that Accuracy can reward prediction of only the dominant-class in imbalanced datasets \cite{he2009learning}, potentially resulting in no positive predictions and an $F_1$-Score of 0, as in line (3) of Table \ref{tbl:direct-results} for the Mammography and Kaggle datasets.


\textit{AUROC computed with soft sets.}
Neither optimizing AUROC using our method nor optimizing AUROC via the WCW statistic \cite{yan2003optimizing} consistently outperform traditional BCE.
In Table \ref{tbl:direct-results}, lines (4) and (5) have lower $F_1$-Score, Accuracy, and AUROC than line (6).
This may be due in part to the AUROC metric's challenge with scale assumptions \cite{flach2015precision}.
Moreover, AUROC is not linear-fractional \cite{koyejo2014consistent}, unlike $F_\beta$-Score and Accuracy. Therefore, AUROC has no unbiased estimator of gradient direction \cite{bao2020calibrated}, which may also contribute to lackluster performance.

Note that in comparison to \cite{fathony2020ap}, the adversarial approach $F_1\dagger$ did not perform as well in our experiments.
The difference could be due to different datasets, and because we used a PyTorch implementation provided by the author while \cite{fathony2020ap} utilized Julia and Flux ML.
Training networks for the tabular datasets on $F_1$-Score using our method on a CPU took a median time of $2.3 \pm 0.16$ minutes whereas the adversarial approach on CPU took a median time of $70 \pm 118$ minutes.

\subsection{Experiments on Image Data}
\label{sec:direct-optimization-experiment-image}

\begin{table}
  \caption{Losses (rows): $F_1$ and Accuracy and via the proposed method. Bold indicates performance better than or equal to BCE baseline.
  }
  \label{tbl:results-cifar}
  \centering
{\small
\begin{tabular}{lrcccc}
\toprule
\multicolumn{2}{c}{} & \multicolumn{2}{c}{\textbf{CIFAR-10-Transportation} ($\mu\pm\sigma$)} & \multicolumn{2}{c}{\textbf{CIFAR-10-Frog} ($\mu\pm\sigma$)} \\
\cmidrule(lr){3-4} \cmidrule(lr){5-6}
& Loss & $\text{F}_1$-Score & Accuracy&$\text{F}_1$-Score & Accuracy \\ \midrule
(1) & $\text{F}_1*$ & $\mathbf{0.91 \pm 0.00}$ & $\mathbf{0.93 \pm 0.00}$ & $\mathbf{0.73 \pm 0.01}$ & $\mathbf{0.95 \pm 0.00}$ \\
(2) & $\text{Accuracy}*$ & $\mathbf{0.92 \pm 0.00}$ & $\mathbf{0.93 \pm 0.00}$ & $\mathbf{0.74 \pm 0.01}$ & $\mathbf{0.95 \pm 0.00}$ \\
(3) & BCE & $0.88 \pm 0.01$ & $0.91 \pm 0.00$ & $0.59 \pm 0.04$ & $0.94 \pm 0.00$ \\
\end{tabular}
}
\end{table}

We conducted experiments similar to the those in Sec. \ref{sec:direct-optimization-experiment} with higher dimensional data using two different binary image datasets created from the CIFAR-10 dataset \cite{krizhevsky2009learning}. 
The CIFAR-10-Transportation and CIFAR-10-Frog datasets had a 40\% and 10\% positive class balance, respectively.

We focused on Accuracy and $F_1$-Score, given the limitations with AUROC found in earlier experiments.
We also excluded the adversarial approach to optimize $F_1$-Score \cite{fathony2020ap} due to long runtime on tabular data.
Otherwise, we use the same set up from Sec. \ref{sec:direct-optimization-experiment}.

Overall, our findings with image data in Table \ref{tbl:results-cifar} are consistent with tabular results from Sec. \ref{sec:direct-optimization-experiment}.
%
Our method of optimizing $F_1$-Score over the soft-set confusion matrix performed better than traditional BCE for both datasets when evaluated on both $F_1$-Score and Accuracy. In Table \ref{tbl:results-cifar}, line (1) has higher scores than line (3).
%
Results from our method of optimizing Accuracy over the soft-set confusion matrix (line (2) in Table \ref{tbl:results-cifar}) are better than the results for BCE (line (3) in Table \ref{tbl:results-cifar}).


\subsection{Balancing Between Precision and Recall}
\label{sec:precision-recall}
Our method also allows training-time optimization that balances between precision and recall using $F_\beta$-Score. This approach to training is particularly useful in real-world scenarios where there is a high cost associated with missed detections.
This type of metric is difficult to optimize effectively for using a typical BCE loss because BCE is not aware of any preference towards precision or recall.

Results in Table \ref{tbl:fb-results} show that using the proposed method to optimize $F_{\beta}$-Score is an effective way of maintaining maximum classifier performance while balancing between precision and recall at a ratio appropriate for a given task. Increasing values of $\beta$ correspond to an increased preference toward recall with small loss of total performance measured by $F_1$-Score. 
Similar results on optimizing for $F_\beta$-Score with additional datasets are provided in Sec. \ref{supp:sec:precision-recall} of the Supplementary Material.

\begin{table}
  \caption{\textbf{Mammography} ($\mu\pm\sigma$) dataset: $F_\beta$ ($\beta = \{1,2,3\}$) loss using the proposed method, to balance between precision and recall while maximizing $F_1$-Score.}
  \label{tbl:fb-results}
  \centering
{\small
\begin{tabular}{lrccccc}
\toprule
& Loss & $F_1$-Score & $F_2$-Score & $F_3$-Score & Precision & Recall \\ \midrule
(1) & $\text{F}_1*$ & $0.61 \pm 0.06$ & $0.57 \pm 0.06$ & $0.56 \pm 0.06$ & $0.70 \pm 0.06$ & $0.55 \pm 0.07$ \\
(2) & $\text{F}_2*$ & $0.63 \pm 0.04$ & $0.67 \pm 0.04$ & $0.69 \pm 0.04$ & $0.57 \pm 0.05$ & $0.71 \pm 0.04$ \\
(3) & $\text{F}_3*$ & $0.57 \pm 0.03$ & $0.69 \pm 0.02$ & $0.75 \pm 0.02$ & $0.44 \pm 0.04$ & $0.81 \pm 0.03$ \\
\end{tabular}
}
\end{table}

\section{Related work}


Our work is inspired by research on the direct optimization of evaluation metrics for binary classification.
This includes plug-in methods that empirically estimate a threshold for a classifier on a metric. For example,  \cite{narasimhan2014statistical} demonstrated the applicability of plug-in classifiers to optimize $F_1$-Score with linear models. For metrics based on linear combinations of confusion matrix set cardinalities, \cite{koyejo2014consistent} identified an optimal plug-in classifier with a metric-dependent threshold. Also, \cite{kar2015surrogate} explored the Precision@K metric for linear models in the context of ranking.
Our approach is not a competitor to plug-in methods, but rather an approach to train a neural network classifier on a differentiable approximation of a metric based on the confusion matrix. As such, it could be used in conjunction with a plug-in method, if desired.

Works such as \cite{kar2014online, busa2015online} optimized specific metrics like Precision@K and $F_1$-Score in online learning, which is characterized by the sequential availability of data. Our work does not address online learning, but batch learning methods.
Additionally, other work has focused on optimizing AUROC \cite{yan2003optimizing, herschtal2004optimising}, F-score \cite{dembczynski2011exact, zhao2013beyond}, and AUPRC in the context of ranking \cite{eban2016scalable}.
Rather than focusing on a single specific metric, we provide a flexible method for optimizing a neural network using approximations of varied metrics based on the confusion matrix.
For example, our experiments in Sec. \ref{sec:precision-recall} show how tradeoffs between precision and recall can be made by adjusting the $\beta$ parameter of the $F_\beta$-score during training with our method.


In the field of computer vision, differentiable surrogate losses have been proposed for the $F_1$-Score \cite{chamidullin2021deep, bozcan2021gridnet}, Jaccard Index \cite{berman2018lovasz, rahman2016optimizing, rezatofighi2019generalized}, and Dice score \cite{nordstrom2020calibrated,sudre2017generalised,milletari2016v,bertels2019optimizing}.
However, since these methods are applied to difficult problems in computer vision, they incorporate a particular surrogate, such as a surrogate for the $F_1$-Score, into a larger composition of losses.
In our experiments, we instead focus on evaluating the impact of individual losses approximating a desired evaluation metric. We also evaluate the theoretical merits of optimizing a differentiable surrogate for confusion-matrix based metrics using an approximation of the Heaviside step function and soft-set confusion matrix values.

Recently, adversarial approaches have emerged as another related area of research. Wang et al. \cite{wang2015adversarial} used a structured support vector machine and reported performance on Precision@K as well as $F_1$-Score.
Fathony and Kolter \cite{fathony2020ap}, which we compare against, improved performance via a marginalization technique that ensured polynomial time convergence. The authors evaluated their adversarial approach on Accuracy and $F_1$-Score, among other metrics, while reporting performance relative to BCE \cite{fathony2020ap}. Downsides of the latter approach are that it is limited to a small batch size, on the order of 25 samples, and has cubic runtime complexity.
Our approach does not limit batch sizes and has a worst-case runtime complexity equivalent to the runtime complexity of the confusion-matrix based metric being used as a loss.


\section{Broader impact and ethics}
\label{sec:ethics}
Our work was motivated by the potentially wide ranging impact that more flexible and robust binary classifiers could have across application domains such as social group dynamics, sociology, and economics, which we explored in our experiments.
Improving the tools used in these areas of research has the potential to positively impact human quality of life, but are still susceptible to data bias.
These tools should be used with care when building safe and responsible artificial intelligence systems.

\section{Conclusion}
\label{sec:conclusion}

We proposed a novel method to optimize for confusion-matrix based metrics by using a Heaviside function approximation and soft-set membership.
Our method addresses the common training-evaluation gap when working with binary neural network classifiers: these networks are typically trained with BCE loss, but evaluated using a confusion-matrix based metric, such as $F_1$-Score.

Our theoretical analyses showed that soft-set confusion matrix based metrics, such as $F_1^s$, are Lipschitz continuous and are likely to converge in expectation to the true metric's expectation.
Additionally, our experiments showed the feasibility of using our method to optimize confusion-matrix based metrics.
While many factors play into final classifier performance, such as balance of samples in the dataset, the desired evaluation metric, the approximation of H, and network hyperparameters, we found that in many cases, final classifier performance can be improved by bridging the training-evaluation gap using our method.
In fact, optimizing model parameters for $F_1$ with soft sets resulted in better evaluation results with $F_1$-Score and better or comparable evaluation results with Accuracy and AUROC. 
Our approach also outperformed other methods that directly optimize for a specific metric.
For example, Fathony and Kolter's \cite{fathony2020ap} adversarial approach applied to $F_1$-Score had much longer training time and resulted in worse performance than our approach.
In the future, we are interested in further studying how our method can be applied to multi-class classification problems.




\setcounter{section}{0}

\title{Supplementary Material \\ Bridging the Gap: Unifying the Training and Evaluation of Neural Network Binary Classifiers}

\maketitle

\section{Theoretical Grounding}
\label{supp:sec:theory}
This section provides the proofs mentioned in Section \ref{sec:theory} of the main paper.

\subsection{Lipschitz Continuity of Metrics Based on Soft-Set Confusion Matrix Values}
\label{supp:sec:theory:lipschitz}


\newcounter{realsection}
\setcounter{realsection}{\value{section}}
\setcounter{section}{4}
\begin{theorem}
The linear Heaviside function approximation $\mathcal{H}^l$ is Lipschitz continuous with Lipschitz constant $M = \max \{m_1, m_2, m_3\}$. 
\end{theorem}
\setcounter{section}{\value{realsection}}
\begin{proof}
Recall that $\mathcal{H}^l$ is piecewise linear, consisting of three line segments of slopes:

{\small
    \noindent\begin{minipage}{.33\linewidth}
    \centering
        $$
        m_1 = \frac{\delta}{\tau-\frac{\tau_m}{2}}
        $$
    \end{minipage}%
    \noindent\begin{minipage}{.33\linewidth}
    \centering
        $$
        m_2 = \frac{1-2\delta}{\tau_m}
        $$
    \end{minipage}%
    \noindent\begin{minipage}{.33\linewidth}
    \centering
        $$
        m_3 = \frac{\delta}{1-\tau-\frac{\tau_m}{2}}
        $$
    \end{minipage}
}

where $\tau_m = \min \{\tau, 1 - \tau \}$. 

Consider any fixed $\tau$. $\mathcal{H}^l$ is continuous, because each piecewise linear component is continuous, and we can computationally verify that $\mathcal{H}^l(p, \tau)$ is defined to be continuous at the two points $p = \tau \pm \frac{\tau_m}{2}$. Let $M = \max \{m_1, m_2, m_3 \}$. Then, we show that the slope of any secant line must be nonnegative and bounded by $M$.

For simplicity, let $f(p) = \mathcal{H}^l(p, \tau)$. Consider any two points $0 \le p_1, p_2 \le 1$, and assume without loss of generality that $p_1 \le p_2$. If $p_1, p_2 \le \tau - \frac{\tau_m}{2}$, then $|f(p_2) - f(p_1)| = m_1 |p_2 - p_1|$. If $\tau - \frac{\tau_m }{2}\le p_1, p_2 \le \tau + \frac{\tau_m}{2}$, then $|f(p_2) - f(p_1)| = m_2 |p_2 - p_1|$. Furthermore, if $\tau + \frac{\tau_m }{2}\le p_1, p_2$, then $|f(p_2) - f(p_1)| = m_3 |p_2 - p_1|$. In all three of these cases, $|f(p_2) - f(p_1)| \le M |p_2 - p_1|$.

Otherwise, if $p_1$ and $p_2$ do not both lie within the bounds of same singular line segment, then there are three more cases. If $p_1 \le \tau - \frac{\tau_m}{2} \le p_2 \le \tau + \frac{\tau_m}{2}$, then since $p_1 \le p_2$ and $f$ is nondecreasing:
{\small
\begin{equation*}
\begin{alignedat}{2}
    |f(p_2) - f(p_1)| &= \left|f(p_2) - f\left(\tau - \frac{\tau_m}{2}\right)\right| + \left|f\left(\tau - \frac{\tau_m}{2}\right) - f(p_1)\right| \\
    &= m_2 \left|p_2 - \left(\tau - \frac{\tau_m}{2}\right)\right| + m_1 \left|\left(\tau - \frac{\tau_m}{2}\right) - p_1\right| \\
    &\le M|p_2 - p_1|
\end{alignedat}
\end{equation*}
}

Similarly, if $\tau- \frac{\tau_m}{2} \le p_1 \le \tau + \frac{\tau_m}{2} \le p_2$, then: 
{\small
\begin{equation*}
\begin{alignedat}{2}
    |f(p_2) - f(p_1)| &= \left|f(p_2) - f\left(\tau + \frac{\tau_m}{2}\right)\right| + \left|f\left(\tau + \frac{\tau_m}{2}\right) - f(p_1)\right| \\
    &= m_3 \left|p_2 - \left(\tau + \frac{\tau_m}{2}\right)\right| + m_2 \left|\left(\tau + \frac{\tau_m}{2}\right) - p_1\right| \\
    &\le M|p_2 - p_1|
\end{alignedat}
\end{equation*}
}

Finally, if $p_1 \le \tau - \frac{\tau_m}{2}$ and $p_2 \ge \tau + \frac{\tau_m}{2}$, then: 
{\small
\begin{equation*}
\begin{alignedat}{2}
    |f(p_2) - f(p_1)| &= \left|f(p_2) - f\left(\tau + \frac{\tau_m}{2}\right)\right| + \left|f\left(\tau + \frac{\tau_m}{2}\right) - f\left(\tau - \frac{\tau_m}{2}\right)\right| + \left|f\left(\tau - \frac{\tau_m}{2}\right) - f(p_1)\right| \\
    &= m_3 \left|p_2 - \left(\tau + \frac{\tau_m}{2}\right)\right| + m_2\left| \left(\tau + \frac{\tau_m}{2}\right) - \left(\tau - \frac{\tau_m}{2}\right)\right| + m_1\left|\left(\tau - \frac{\tau_m}{2}\right) - p_1\right| \\
    &\le M|p_2 - p_1|
\end{alignedat}
\end{equation*}
}

This exhausts all cases, and so for all $p_1, p_2$, we have $|\mathcal{H}^l(p_2, \tau) - \mathcal{H}^l(p_1, \tau)| = |f(p_2) - f(p_1)| \le M|p_2 - p_1|$. Thus, $\mathcal{H}^l$ is Lipschitz continuous with Lipschitz constant $M$. 
\end{proof}


\setcounter{realsection}{\value{section}}
\setcounter{section}{4}
\begin{theorem}
Every entry of the soft-sets confusion matrix based on the Heaviside approximations are Lipschitz continuous in the output of a neural network.
\end{theorem}
\setcounter{section}{\value{realsection}}
\begin{proof}
Without loss of generality, we prove that the $|TP_s|$ entry is Lipschitz continuous in the input vector $p = (p_1, \ldots, p_n)$ corresponding to outputs of the neural network given inputs $x = (x_1, \ldots, x_n)$ and labels $y = (y_1, \ldots, y_n)$. The Lipschitz continuity of all the other entries ($|FN_s|$, $|FP_s|$, and $|TN_s|$) follows similarly by symmetry. For any sample tuple $(p_i, y_i, \tau)$, recall that:
{\small
\begin{equation*}
\begin{alignedat}{2}
\mathit{tp_s}(p_i, y_i, \tau) &= 
  \begin{cases}
    \mathcal{H}^l(p_i,\tau) & y_i = 1 \\
    0 & \text{otherwise}
  \end{cases} 
\end{alignedat}
\end{equation*}
}

where the final value $|TP_s|$ is calculated from the summation $|TP_s| = \sum_{i=1}^n\mathit{tp_s}(p_i, y_i, \tau)$ over all sample tuples $(p_i, y_i, \tau)$. Since $\mathit{tp_s}(p_i, y_i, \tau)$ evaluates to $0$ if $y_i \neq 1$, this sum is equivalent to $|TP_s| = \sum_{y_i = 1} \mathcal{H}^l(p_i,\tau)$. 

By Theorem \ref{main:thm:HLip}, we know that $\mathcal{H}^l(p_i, \tau)$ is Lipschitz continuous in $p_i$ over the domain $p_i \in [0, 1]$.
Any function that takes as input a vector is Lipschitz continuous if it is Lipschitz continuous in each entry of its input.
It follows that $|TP_s| = \sum_{y_i = 1} \mathcal{H}^l(p_i, \tau)$ is Lipschitz continuous in the input vector $p$. 

By a similar reasoning, it follows that the other entries $|FN_s|$, $|FP_s|$, and $|TN_s|$ are all also Lipschitz continuous in the input vector $p$.
\end{proof}

\subsection{Approximation of Confusion-Matrix Based Metrics with Soft Sets}
\label{supp:sec:theory:converge}

Following the ideas in Section \ref{sec:theory:convergence} of the main paper, we generalize the proof for $F_1$-Score and show that, under similar assumptions, convergence holds for certain other objective functions. Specifically, our proof holds for all continuous objective functions that can be expressed solely in terms of the ratios of entries of the confusion matrix to $n$, the size of the dataset. Most metrics, including Accuracy, Arithmetic Mean, $F_\beta$-Score, Jaccard, and Geometric Mean all satisfy this property. Our generalized proof assumes a fixed threshold value $\tau$, but we also show that convergence holds for AUROC, which is computed over a range of threshold values.

As before, consider a dataset of size $n$ with $\{x_1, ..., x_n\}$ examples and $\{y_1, ..., y_n\}$ labels. Suppose a network outputs a probability $p_i$ that the label $y_i=1$. Since the specific outputs $p_i$ are unknown and may change across iterations, we again assume that $p_i$ is a random variable. When computing the entries of our confusion matrix, $p_i$ is passed through the Heaviside function $H$, which generates an output $\hat y_i^H = H(p_i, \tau)$, where $\hat y_i^H \in \{0, 1\}$. Then $|TP| = \sum_{y_i=1} \hat y_i^H$ and $|FP| = \sum_{y_i=0} \hat y_i^{H}$. On the other hand, under our proposed approximation, $p_i$ is passed through the Heaviside approximation $\mathcal{H}$, which generates an output $\hat y_i^{\mathcal{H}} = \mathcal{H}(p_i, \tau)$. In this case, $\hat y_i^{\mathcal{H}}$ can take on any real value between $0$ and $1$. Then $|TP_s| = \sum_{y_i=1} \hat y_i^{\mathcal{H}}$ and $|FP_s| = \sum_{y_i=0} \hat y_i^{\mathcal{H}}$ are entries of the soft set confusion matrix. 

\subsubsection{Continuous Generalization and Accuracy}
Consider a network trained on a dataset with $rn$ positive and $(1-r)n$ negative elements, where $r \in [0,1]$ is some constant. Let a loss function $\ell$ be continuous in the ratio of each entry of the confusion matrix to $n$, the size of the dataset. Hence, $\ell$ can be written as a function of $\frac{|TP|}{n}$, $\frac{|FP|}{n}$, $\frac{|TN|}{n}$, and $\frac{|FN|}{n}$. But since $|TP| + |FN| = nr$ and $|FP| + |TN| = (1-r)n$, $\frac{|FN|}{n} = r - \frac{|TP|}{n}$ and $\frac{|TN|}{n} = (1-r) - \frac{|FP|}{n}$. Thus, $\frac{|FN|}{n}$ and $\frac{|TN|}{n}$ can be represented in terms of $\frac{|TP|}{n}$ and $\frac{|FP|}{n}$, so $\ell\left (\frac{|TP|}{n}, \frac{|FP|}{n} \right)$ can be expressed as a continuous function on just $\frac{|TP|}{n}$ and $\frac{|FP|}{n}$. 

Thus, the loss computed using the Heaviside step function can be expressed as: 
{\small
\begin{equation}
\label{eq:ell-yhat}
\begin{alignedat}{2}
 \ell = \ell\left(\frac{1}{n}\sum_{y_i = 1} \hat y_i^H, \frac{1}{n}\sum_{y_i = 0} \hat y_i^H \right)
\end{alignedat}
\end{equation}
}

The loss when computed from soft sets is then:
{\small
\begin{equation}
\label{eq:ells-yhat}
\begin{alignedat}{2}
 \ell^s = \ell \left ( \frac{1}{n}\sum_{y_i = 1} \hat y_i^{\mathcal{H}}, \frac{1}{n}\sum_{y_i = 0} \hat y_i^{\mathcal{H}} \right )
\end{alignedat}
\end{equation}
}

As in Section \ref{sec:theory:convergence} of the main paper, suppose a classifier correctly classifies any positive example as a true positive with probability $u$ and any negative example as a false positive with probability $v$.
Also, assume that all classifications are independent.
Because the loss function $\ell$ is calculated with discrete $\hat y_i^H$, we assume that the classifier will classify examples as a random variable $\hat y_i^H \sim \Bern(uy_i + v(1-y_i))$. 
Thus, $\hat y_i^H \sim \Bern(u)$ if $y_i=1$, and $\hat y_i^H \sim \Bern(v)$ if $y_i=0$. 

Since $\ell^s$ can take on continuous values in $[0, 1]$, we consider that $\hat y_i^{\mathcal{H}}$ is a random variable drawn from a Beta distribution, which has support $[0, 1]$. In particular, assume $\hat y_i^{\mathcal{H}} \sim \Beta(\alpha_uy_i + \alpha_v(1-y_i), \beta_uy_i + \beta_v(1-y_i))$. Hence, $\hat y_i^{\mathcal{H}} \sim \Beta(\alpha_u, \beta_u)$ if $y_i=1$, and $\hat y_i^{\mathcal{H}} \sim \Beta(\alpha_v, \beta_v)$ if $y_i=0$. Let $\frac{\alpha_u}{\alpha_u + \beta_u} = u$ and $\frac{\alpha_v}{\alpha_v+\beta_v} = v$, so for any $i$, $\E\left[\hat y_i^{\mathcal{H}}\right]=u$ if $y_i = 1$, and $\E\left[\hat y_i^{\mathcal{H}}\right]=v$ if $y_i = 0$.

Under the above assumptions, both $\ell$ and $\ell^s$ have the same average classification correctness: for any given $i$, $\E\left[\hat y_i^H\right] = \E\left[\hat y_i^{\mathcal{H}}\right]$. Also, there exists some $\alpha_u, \beta_u, \alpha_v, \beta_v$ such that the distributions of $\hat y_i^H = H(p_i, \tau)$ and $\hat y_i^{\mathcal{H}} = \mathcal{H}(p_i, \tau)$ can both hold simultaneously under the same network for all $i$.

If we let $\sum_{y_i = 1} \hat y_i^H = U \sim \Bin(nr, u)$ and $\sum_{y_i = 0} \hat y_i^H = V \sim \Bin(n(1-r), v)$ be the independent random variables denoting the number of true positives and false positives in a sequence of $n$ independent predictions, then by the Strong Law of Large Numbers, 
$\frac{1}{nr}U \asconv u$ and $\frac{1}{n(1-r)}V \asconv v$ both converge with probability $1$ as $n \to \infty$.
Hence, $\frac{U}{n} \asconv ru$ and $\frac{V}{n} \asconv (1-r)v$.
We therefore have, from the Continuous Mapping Theorem, that as $n \to \infty$:
{\small
\begin{equation}
\begin{alignedat}{2}
 \ell = \ell\left(\frac{1}{n}\sum_{y_i = 1} \hat y_i^H, \frac{1}{n}\sum_{y_i = 0} \hat y_i^H \right) = \ell\left(\frac{U}{n}, \frac{V}{n} \right) \asconv \ell(ru, (1-r)v)
\end{alignedat}
\end{equation}
}

For $\ell^s$, let $U^s=\sum_{y_i = 1} \hat y_i^{\mathcal{H}}$ and $V^s = \sum_{y_i = 0} \hat y_i^{\mathcal{H}}$ be the independent random variables denoting the total amount of true positives and false positives in the soft set case. Then, $\ell^s$ from Eq. \eqref{eq:ells-yhat} becomes:
{\small
\begin{equation}
\begin{alignedat}{2}
 \ell^s = \ell \left ( \frac{1}{n}\sum_{y_i = 1} \hat y_i^{\mathcal{H}}, \frac{1}{n}\sum_{y_i = 0} \hat y_i^{\mathcal{H}} \right ) = \ell\left(\frac{U^s}{n}, \frac{V^s}{n}\right)
\end{alignedat}
\end{equation}
}

But by the Strong Law of Large Numbers, $\frac{1}{nr}U^s \asconv \frac{\alpha_u}{\alpha_u + \beta_u} = u$. Similarly, $\frac{1}{n(1-r)}V^s \asconv \frac{\alpha_v}{\alpha_v+\beta_v}=v$ also converges with probability $1$ as $n \to \infty$. Thus, $\frac{U^s}{n} \asconv ru$ and $\frac{V^s}{n} \asconv (1-r)v$. By the Continuous Mapping Theorem:
{\small
\begin{equation}
\begin{alignedat}{2}
 \ell^s = \ell\left(\frac{U^s}{n}, \frac{V^s}{n}\right) \asconv \ell (ru, (1-r)v)
\end{alignedat}
\end{equation}
}

Thus, $\ell$ and $\ell^s$ both converge almost surely to the same value as $n \to \infty$. 
Since $\ell(ru, (1-r)v)$ is a finite constant, $\E[\ell], \E[\ell^s] \to \ell(ru, (1-r)v)$ by the Bounded Convergence Theorem. This means that the $\ell^s$ value is an asymptotically unbiased estimator for the expected loss $\ell$, and we expect average loss values to converge to $\ell^s$ as $n \to \infty$, under our setup.

Since $\text{Accuracy} = \frac{|TP| + |TN|}{n}$ is a continuous function on $\frac{|TP|}{n}$ and $\frac{|TN|}{n}$, it follows that Accuracy computed from soft-sets, $\text{Accuracy}^s$, and Accuracy computed using the Heaviside step function both converge almost surely to the same value under this setup as $n \to \infty$. Similarly, $\E [\text{Accuracy}^s]$ and $\E [\text{Accuracy}]$ both converge to the same expectation as $n \to \infty$.

\subsubsection{AUROC}

Let the true positive rate, $\text{TPR}(\tau)$, and false positive rate, $\text{FPR}(\tau)$, be functions of the threshold value $\tau$. Then AUROC is defined as $\text{AUROC}=\int_{x=0}^1 \text{TPR}(\text{FPR}^{-1}(x)) \, dx$. In our case of binary classification, the choice of threshold value $\tau$ can range from $0$ to $1$.

Consider a network trained on a dataset with $rn$ positive and $(1-r)n$ negative elements, where $r \in [0,1]$ is some constant. Previously, we treated the probabilities $u$ and $v$ as constants since $\tau$ was fixed. Now, assume that $u = u(\tau)$ and $v = v(\tau)$ are both parameterized by $\tau$. Both $u$ and $v$ must be nonincreasing in $\tau$ because increasing the threshold value will never increase the number of positives classified. Furthermore, we must have $u(0) = v(0) = 1$ and $u(1) = v(1) = 0$ as the boundary conditions. Since $\tau$ is no longer fixed, we also let $\hat y_i^H(\tau) = H(p_i, \tau)$ and $\hat y_i^{\mathcal{H}}(\tau) = \mathcal{H}(p_i, \tau)$ both be parameterized by $\tau$ and be nonincreasing in $\tau$.  

In practice, since binary classification has discrete true positive and false positive rates, we approximate/calculate AUROC by choosing threshold values $0 = \tau_0 < \tau_1 < \ldots < \tau_k = 1$ where $\tau_i = \frac{i}{k}$ for all $0 \le i \le k$ and then using the trapezoidal rule to approximate the area under the ROC curve with the intervals defined by the points $1 = \text{FPR}(\tau_0) \ge \ldots \ge \text{FPR}(\tau_k) = 0$. Since the true positive rate and false positive rate are both monotone in $\tau$ with $\text{TPR}(0) = \text{FPR}(0) = 1$ and $\text{TPR}(1) = \text{FPR}(1) = 0$, the length of the intervals all approach $0$, and so this approximation using the trapezoidal rule converges to the AUROC value as $k \to \infty$. 

Consider some fixed value $k$ and let $0 = \tau_0 < \tau_1 < \ldots < \tau_k = 1$ where $\tau_i = \frac{i}{k}$. Then under this approximation with the trapezoidal rule, we have that the AUROC is: 

{\small
\begin{equation}
\begin{alignedat}{2}
 \text{AUROC} = \frac{1}{2}\sum_{i=1}^k (\text{FPR}(\tau_{i-1}) - \text{FPR}(\tau_i))(\text{TPR}(\tau_i) + \text{TPR}(\tau_{i-1}))
\end{alignedat}
\end{equation}
}


For true AUROC, note that as before that for any given $\tau$, $\text{TPR}(\tau) = \frac{1}{nr}\sum_{y_i=1} \hat y_i^H(\tau)$. Because this is calculated with discrete $\hat y_i^H$, we assume that the classifier will classify examples as a random variable $\hat y_i^H(\tau) \sim \Bern(u(\tau)\cdot y_i + v(\tau)\cdot (1-y_i))$. Thus, $\hat y_i^H(\tau) \sim \Bern(u(\tau))$ if $y_i = 1$ and $\hat y_i^H(\tau) \sim \Bern(v(\tau))$ if $y_i = 0$. Hence, for any $\tau$, $\text{TPR}(\tau) = \frac{1}{nr} \sum_{y_i = 1} \hat y_i^H(\tau) \sim \frac{1}{nr}\Bin(nr, u(\tau))$ is its marginal distribution. Similarly $\text{FPR}(\tau) = \frac{1}{n(1-r)}\sum_{y_i = 0} \hat y_i^H(\tau) \sim \frac{1}{n(1-r)}\Bin(n(1-r), v(\tau))$ for all $\tau$. 

However, both $\text{TPR}(\tau)$ and $\text{FPR}(\tau)$ must be nonincreasing functions in $\tau$. To enforce monotonicity so that $\text{TPR}(\tau_0) \ge \text{TPR}(\tau_1) \ge \cdots \ge \text{TPR}(\tau_k)$ and $\text{FPR}(\tau_0) \ge \cdots \ge \text{FPR}(\tau_k)$, we construct their joint distribution as follows. First, consider $n$ uniformly distributed i.i.d. random variables $X_1, \ldots, X_n \sim \Unif(0, 1)$. Then for all $i$ and all $0 \le j \le k$, we let:

{\small
\begin{equation}
\begin{alignedat}{2}
 \hat y_i^H(\tau_j) = \begin{cases} 1 & X_i \le u(\tau_j)\cdot y_i + v(\tau_j)\cdot (1-y_i) \\ 0 & X_i > u(\tau_j)\cdot y_i + v(\tau_j)\cdot (1-y_i) \end{cases}
\end{alignedat}
\end{equation}
}

Indeed, under this construction, the marginal distributions $\text{TPR}(\tau_j) \sim \frac{1}{nr} \Bin(nr, u(\tau_j))$ and $\text{FPR}(\tau_j) \sim \frac{1}{n(1-r)} \Bin(n(1-r), v(\tau_j))$ both hold. We now, however, also have that $\text{TPR}(\tau_0) \ge \cdots \ge \text{TPR}(\tau_k)$ and $\text{FPR}(\tau_0) \ge \cdots \ge \text{FPR}(\tau_k)$.

Since each marginal distribution $\text{TPR}(\tau_i) \sim \frac{1}{nr}\Bin(nr, u(\tau_i))$, by the Law of Large Numbers, for all $0 \le i \le k$, $\text{TPR}(\tau_i) \asconv u(\tau_i)$. Similarly, $\text{FPR}(\tau_i) \asconv v(\tau_i)$. Thus, by the Continuous Mapping Theorem, the true AUROC is:
{\small
\begin{equation}
\begin{alignedat}{2}
 \text{AUROC} &= \frac{1}{2}\sum_{i=1}^k (\text{FPR}(\tau_{i-1}) - \text{FPR}(\tau_i))(\text{TPR}(\tau_i) + \text{TPR}(\tau_{i-1})) \\
    &\asconv \frac{1}{2} \sum_{i=1}^k (v(\tau_{i-1}) - v(\tau_i))\cdot (u(\tau_i) + u(\tau_{i-1}))
\end{alignedat}
\end{equation}
}

Now, consider AUROC computed from soft sets (which we denote as $\text{AUROC}^s$). $\text{AUROC}^s$ can take on continuous values in $[0, 1]$, so we consider that $\hat y_i^{\mathcal{H}}$ is a random variable drawn from a Beta distribution, which has support $[0, 1]$. In particular, for any $\tau$, assume $\hat y_i^{\mathcal{H}}(\tau) \sim \Beta(\alpha_u(\tau)\cdot y_i + \alpha_v(\tau)\cdot (1-y_i), \beta_u(\tau)\cdot y_i + \beta_v(\tau)\cdot (1-y_i))$. Hence, $\hat y_i^{\mathcal{H}} \sim \Beta(\alpha_u(\tau), \beta_u(\tau))$ if $y_i=1$, and $\hat y_i^{\mathcal{H}} \sim \Beta(\alpha_v(\tau), \beta_v(\tau))$ if $y_i=0$. Let $\frac{\alpha_u(\tau)}{\alpha_u(\tau) + \beta_u(\tau)} = u(\tau)$ and $\frac{\alpha_v(\tau)}{\alpha_v(\tau)+\beta_v(\tau)} = v(\tau)$, so for any $i$, $\E\left[\hat y_i^{\mathcal{H}}(\tau)\right]=u(\tau)$ if $y_i = 1$, and $\E\left[\hat y_i^{\mathcal{H}}(\tau)\right]=v(\tau)$ if $y_i = 0$.

Once again, we show that monotonicity can be enforced across the marginal distributions for corresponding true positive and false positive rates computed from soft sets: $\text{TPR}^s(\tau) = \frac{1}{nr}\sum_{y_i = 1} \hat y_i^{\mathcal{H}}(\tau)$ and $\text{FPR}^s(\tau) = \frac{1}{n(1-r)}\sum_{y_i = 0} \hat y_i^{\mathcal{H}}(\tau)$ among the threshold values $\tau_0, \ldots, \tau_k$. It is well known that for any constants $\alpha, \beta, \beta' > 0$ with $\beta > \beta'$, the distribution $\Beta(\alpha, \beta')$ stochastically dominates $\Beta(\alpha, \beta)$: specifically that the cumulative distribution function of $\Beta(\alpha, \beta)$ lies above the cumulative distribution function of $\Beta(\alpha, \beta')$ at every point $x \in (0, 1)$. Let $\alpha_u(\tau) = \alpha_u$ and $\alpha_v(\tau) = \alpha_v$ be constant, with $\beta_u(\tau)$ and $\beta_v(\tau)$ nondecreasing in $\tau$ such that $\frac{\alpha_u}{\alpha_u + \beta_u(\tau)} = u(\tau)$ and $\frac{\alpha_v}{\alpha_v + \beta_v(\tau)} = v(\tau)$. Then, for all $i$, the distribution $\Beta(\alpha_u\cdot y_i + \alpha_v\cdot (1-y_i), \beta_u(\tau')\cdot y_i + \beta_v(\tau')\cdot (1-y_i))$ stochastially dominates $\Beta(\alpha_u\cdot y_i + \alpha_v\cdot (1-y_i), \beta_u(\tau)\cdot y_i + \beta_v(\tau)\cdot (1-y_i))$ if $\tau > \tau'$. We can therefore construct a coupling among the distributions together by considering i.i.d. random variables $X_1', \ldots, X_n'\sim \Unif(0, 1)$ and letting $\hat y_i^{\mathcal{H}}(\tau)$ be the value corresponding to the $X_i'$-th percentile of $\Beta(\alpha_u\cdot y_i + \alpha_v\cdot (1-y_i), \beta_u(\tau)\cdot y_i + \beta_v(\tau)\cdot (1-y_i))$. This construction still maintains the marginal distributions $\hat y_i^{\mathcal{H}} \sim \Beta(\alpha_u, \beta_u(\tau))$ if $y_i=1$ and $\hat y_i^{\mathcal{H}} \sim \Beta(\alpha_v, \beta_v(\tau))$ if $y_i=0$. However, under this construction, $\text{TPR}^s(\tau_0) \ge \cdots \ge \text{TPR}^s(\tau_k)$ and $\text{FPR}^s(\tau_0) \ge \cdots \ge \text{FPR}^s(\tau_k)$ also hold.

By the Law of Large Numbers, for any $0 \le i \le k$, $\text{TPR}^s(\tau_i) = \frac{1}{nr}\sum_{y_i = 1} \hat y_i^{\mathcal{H}}(\tau_i) \asconv \frac{\alpha_u(\tau_i)}{\alpha_u(\tau_i) + \beta_u(\tau_i)} = u(\tau_i)$. Similarly, $\text{FPR}^s(\tau_i) \asconv \frac{\alpha_v(\tau_i)}{\alpha_v(\tau_i) + \beta_v(\tau_i)} = v(\tau_i)$. Thus, by the Continuous Mapping Theorem:
{\small
\begin{equation}
\begin{alignedat}{2}
 \text{AUROC}^s &=\frac{1}{2}\sum_{i=1}^k (\text{FPR}^s(\tau_{i-1}) - \text{FPR}^s(\tau_i))(\text{TPR}^s(\tau_i) + \text{TPR}^s(\tau_{i-1})) \\
    &\asconv \frac{1}{2} \sum_{i=1}^k (v(\tau_{i-1}) - v(\tau_i))\cdot (u(\tau_i) + u(\tau_{i-1}))
\end{alignedat}
\end{equation}
}

Thus, $\text{AUROC}, \text{AUROC}^s \asconv \frac{1}{2} \sum_{i=1}^k (v(\tau_{i-1}) - v(\tau_i))\cdot (u(\tau_i) + u(\tau_{i-1}))$ both converge almost surely to the same value. Since AUROC is bounded between $0$ and $1$, by the Bounded Convergence Theorem, $\E[\text{AUROC}], \E[\text{AUROC}^s] \to \frac{1}{2} \sum_{i=1}^k (v(\tau_{i-1}) - v(\tau_i))\cdot (u(\tau_i) + u(\tau_{i-1}))$.

\section{Heaviside Function Approximation}
\label{supp:sec:heaviside-approx}



\subsection{Sigmoid Approximation: Visual Analysis of Trade-offs when Searching for Optimal $k$}
\label{supp:sec:heaviside-approx-sigmoid}

\begin{figure*}[b!]
    \centering
    \includegraphics[width=\textwidth]{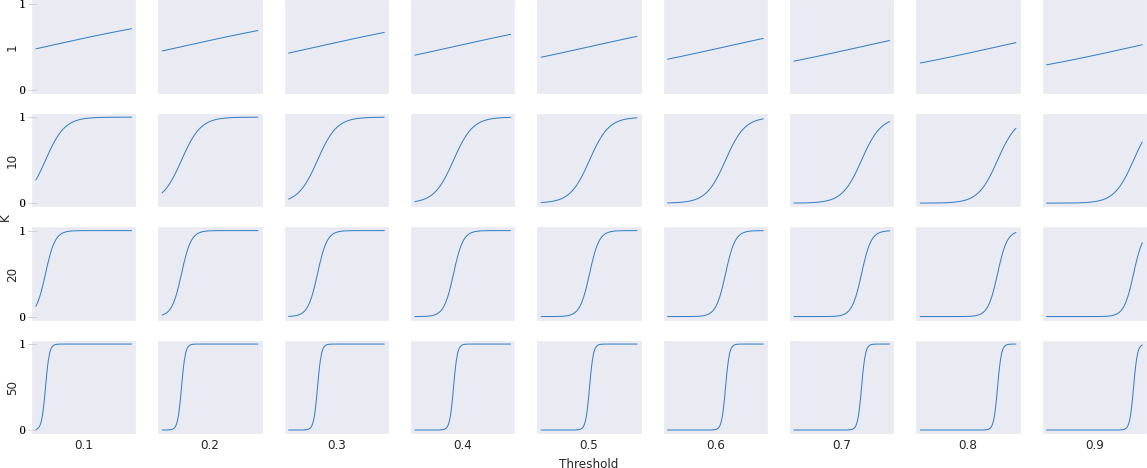}
    \caption{Sigmoid approximation of the Heaviside step function by $k$ and $\tau$. With increased values of $k$, the approximation becomes closer to the Heaviside step function. As $k$ decreases, the number $\tau$ values over which the approximation diverges from the Heaviside step function at the limit increases.}
    \label{fig:sig}
\end{figure*}

\begin{figure*}[b!]
    \centering
    \includegraphics[width=\textwidth]{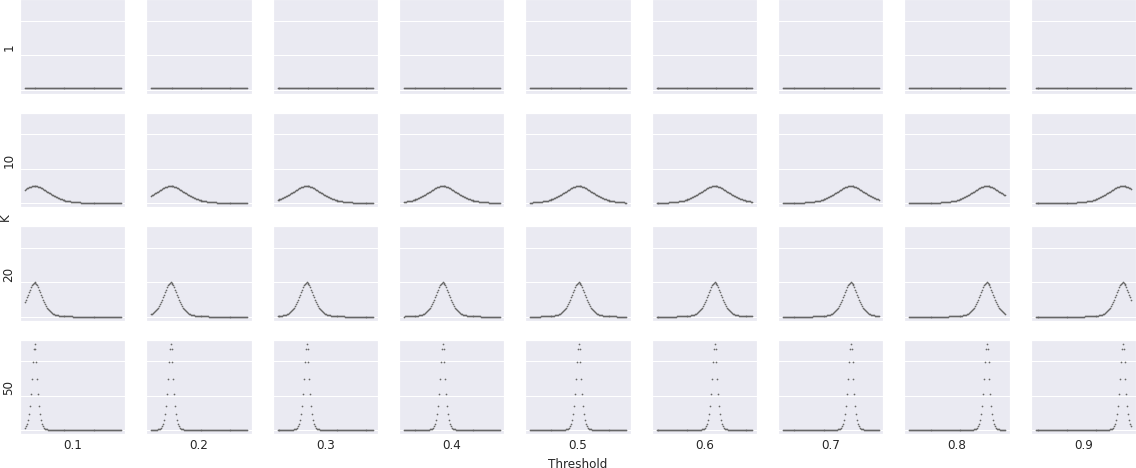}
    \caption{Derivative of the sigmoid approximation of the Heaviside step function by $k$ and $\tau$. As $k$ increases, the range of values with zero gradient increases.}
    \label{fig:sig-grad}
\end{figure*}

As described in Section \ref{sec:heaviside-approximation} of the main paper, a tradeoff must be made when choosing an appropriate value for $k$ in the sigmoid approximation. As $k$ increases, the approximation becomes closer to the Heaviside step function when $\tau = 0.5$, as shown in Figure \ref{fig:sig}. However, the range of values with zero gradient increases, as shown in Figure \ref{fig:sig-grad}.

It is important to note that for soft-set membership calculation, as $k$ decreases, the number of $\tau$ values over which the approximation diverges from the Heaviside step function at the limit increases. In other words, as $k$ decreases, the number of $\tau$ values increases for which the sigmoid approximation does not adhere to the following limits:

\begin{equation} \label{eq:lim}
    \lim_{p \to 0} \mathcal{H}(p,\tau) = 0 \hspace{1em} \forall \, \tau \hspace{2em} \lim_{p \to 1} \mathcal{H}(p,\tau) = 1 \hspace{1em} \forall \, \tau
\end{equation}

Figure \ref{fig:sig} illustrates that for $k=50$, the approximation does approach the limits in Eq. \eqref{eq:lim} for all values of $\tau$.
Contrarily, for $k$=1, the approximation does not approach the limits in Eq. \eqref{eq:lim} for any values of $\tau$.

For our experiments, we searched for the best value of $k$ in $\{1, 10, 20, 50\}$ and found that $k=10$ led to the best performance.

\subsection{Derivation of the Linear Heaviside Approximation}
\label{supp:sec:linear-derivation}

Our proposed linear Heaviside approximation, $\mathcal{H}^l$, is formulated to ensure adherence to the properties described in Section \ref{sec:heaviside-approximation} of the main paper.

We concern ourselves with only the range over [0,1] since we expect the input to represent a probability in [0,1].


For $0 \leq p \leq 1$, we start by specifying the endpoints to ensure Eq. \eqref{eq:lim} above, and we define the point at $p = \tau$ to ensure the property $\mathcal{H}^l(p=\tau,\tau)=0.5$:

\begin{equation} \label{eq:three-points}
    \mathcal{H}^l(p, \tau) =
        \begin{cases}
        0 & \text{if } p = 0 \\
        1 & \text{if } p = 1 \\
        0.5 & \text{if } p = \tau
        \end{cases}
\end{equation}

A natural next step would be to solve for a three point linearly interpolated function using the points defined in Eq. \eqref{eq:three-points}. However, we find that in practice this is difficult to optimize, perhaps due to the fact that the gradient of one segment becomes much greater than the other as $\tau$ approaches $0$ or $1$.
Thus we instead define two more points relative to $\tau$ with the introduction of the parameter $\delta$, which makes optimization via backpropagation more stable. We specify these two points to be equal to $\delta$ and $1-\delta$ when the inputs are halfway the distance between $\tau$ and the nearest endpoint ($0$ or $1$).

Let $\tau_m$ be the distance between $\tau$ and the nearest endpoint of $\mathcal{H}^l$:

\begin{equation}
    \tau_m = \text{min} \{\tau, 1-\tau\}
\end{equation}

Then, we define the two new points at which $\mathcal{H}^l(p, \tau)$ is defined as:

\begin{equation}
    \mathcal{H}^l(p, \tau) =
        \begin{cases}
        \delta & \text{if } p = \tau - \frac{\tau_m}{2} \\
        1-\delta & \text{if } p = \tau + \frac{\tau_m}{2}
        \end{cases}
\end{equation}

\begin{figure}[h!]
    \centering
    \begin{minipage}{0.33\textwidth}
        \centering
        \includegraphics[width=0.95\linewidth]{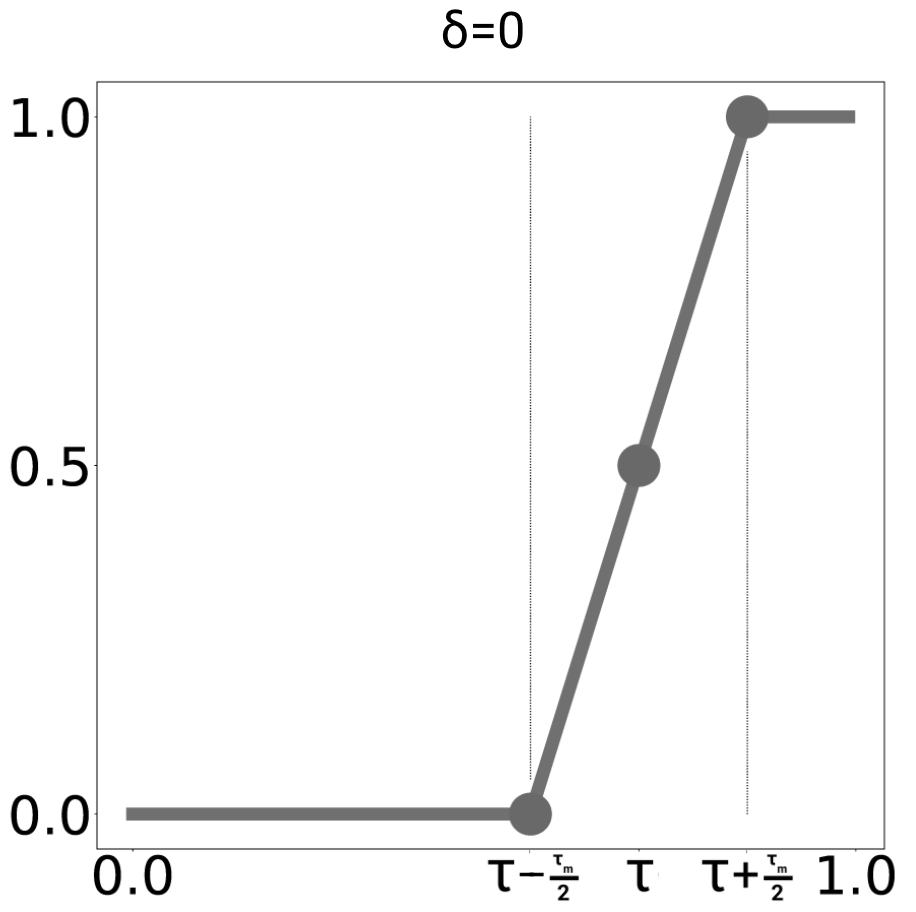}
    \end{minipage}%
    \begin{minipage}{0.33\textwidth}
        \centering
        \includegraphics[width=0.95\linewidth]{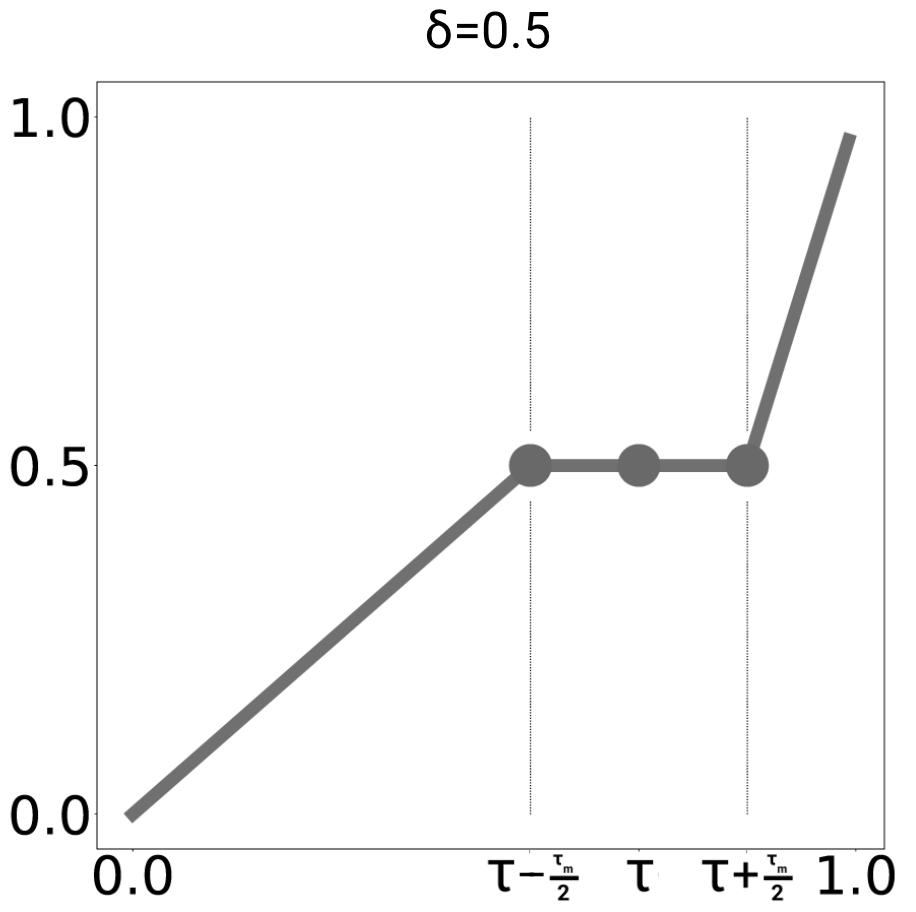}
    \end{minipage}%
    \begin{minipage}{0.33\textwidth}
        \centering
        \includegraphics[width=0.95\linewidth]{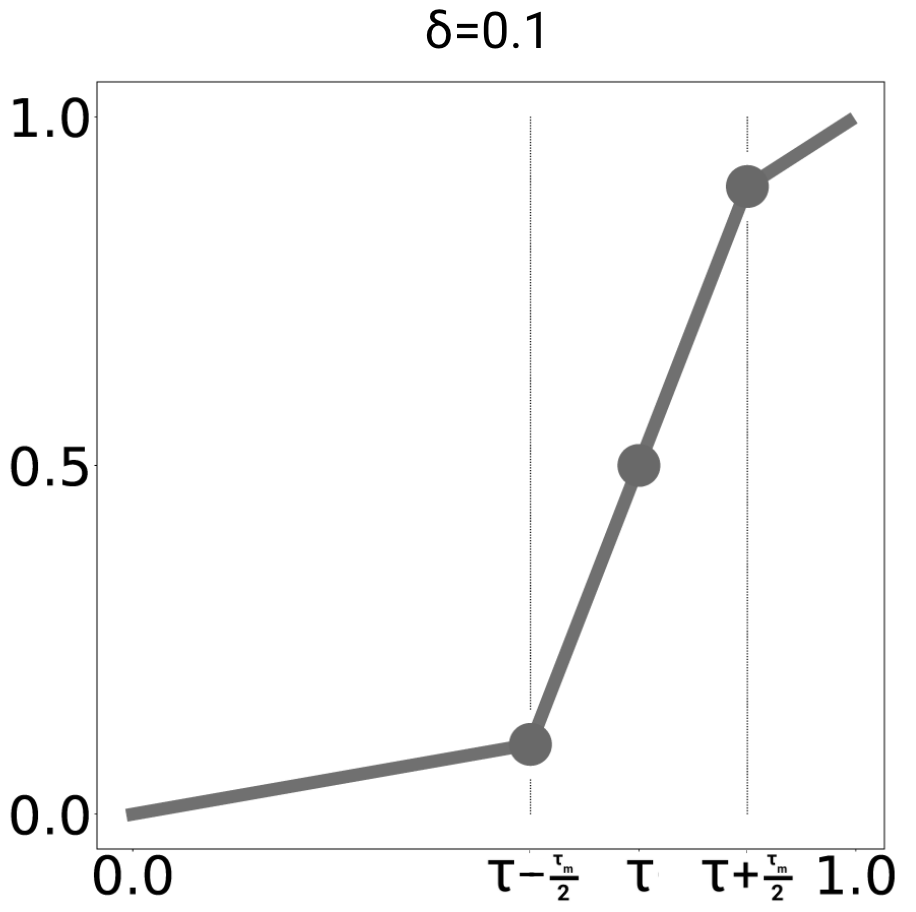}
    \end{minipage}%
    \caption{Linear Heaviside approximation $\mathcal{H}^l$ with $\delta$ values of 0 (left), 0.5 (middle), and 0.1 (right).}
    \label{fig:delta}
\end{figure}

The parameter $\delta$ is chosen such that the function, $\mathcal{H}^l$ remains non-decreasing, limiting the range of $\delta$ to $[0, 0.5]$. Note that, as shown in Figure \ref{fig:delta} (left), a choice of $\delta=0$ would result in gradient equal to zero between $p=0$ and $p=\tau - \frac{\tau_m}{2}$ as well as between $p=\tau + \frac{\tau_m}{2}$ and $p=1$. 
As shown in Figure \ref{fig:delta} (middle), a choice of $\delta=0.5$ would result in a gradient equal to zero between $p=\tau-\frac{\tau_m}{2} $ and $p=\tau + \frac{\tau_m}{2}$. We empirically determined that a value in the range of $0.1 \leq \delta \leq 0.2$ works well in practice. A linear approximation with $\delta=0.1$ is shown in Figure \ref{fig:delta} (right).

In total, the following five points for $\mathcal{H}^l$ have now been defined as follows:

\begin{equation} \label{eq:five-points}
    \mathcal{H}^l(p, \tau) =
        \begin{cases}
        0 & \text{if } p = 0 \\
        \delta & \text{if } p = \tau - \frac{\tau_m}{2} \\
        0.5 & \text{if } p = \tau \\
        1-\delta & \text{if } p = \tau + \frac{\tau_m}{2} \\
        1 & \text{if } p = 1
        \end{cases}
\end{equation}

The property of $\mathcal{H}^l(p=\tau, \tau) = 0.5$ is now satisfied by our formulation. We can use the points to solve for three line segments that fully define $\mathcal{H}^l$. We denote the slope and intercept of each segment from left to right as $m_1, m_2, m_3$ and $b_1, b_2, b_3$ respectively. These slopes, from left to right, are:

\noindent\begin{minipage}{.33\linewidth}
\centering
    $$
    m_1 = \frac{\delta}{\tau-\frac{\tau_m}{2}}
    $$
\end{minipage}%
\noindent\begin{minipage}{.33\linewidth}
\centering
    $$
    m_2 = \frac{1-2\delta}{\tau_m}
    $$
\end{minipage}%
\noindent\begin{minipage}{.33\linewidth}
\centering
    $$
    m_3 = \frac{\delta}{1-\tau-\frac{\tau_m}{2}}
    $$
\end{minipage}


We can solve for the first intercept, $b_1$, with $p=\tau - \frac{\tau_m}{2}$:

\begin{align*}
    \delta &= m_1(\tau - \frac{\tau_m}{2}) + b_1 \\
    b_1 &= \delta - (\frac{\delta}{\tau-\frac{\tau_m}{2}})(\tau - \frac{\tau_m}{2}) \\
    b_1 &= 0
\end{align*}

We then solve for the second intercept, $b_2$, with $p=\tau$:

\begin{align*}
    0.5 &= m_2 \tau + b_2 \\
    b_2 &= 0.5 - m_2 \tau
\end{align*}

Finally, we solve for the third intercept, $b_3$, with $p=\tau + \frac{\tau_m}{2}$:

\begin{align*}
    1-\delta &= m_3 (\tau + \frac{\tau_m}{2}) + b_3 \\
    b_3 &= 1 - \delta - m_3 (\tau + \frac{\tau_m}{2})
\end{align*}

Therefore the linear approximation is:

{\small
\begin{equation}
\mathcal{H}^l =
  \begin{cases}
    p \cdot m_1 & \text{if $p < \tau - \frac{\tau_m}{2}$} \\
    p \cdot m_3 +(1-\delta-m_3(\tau+\frac{\tau_m}{2})) & \text{if $p > \tau + \frac{\tau_m}{2}$} \\
    p \cdot m_2 + (0.5 - m_2\tau) & \text{otherwise}
  \end{cases}
\end{equation}
}

\section{Computational Efficiency}
\label{supp:sec:efficiency}

At training time, an objective function with our proposed method has a runtime linear with regard to the number of samples. This is a large improvement over the adversarial method of Fathony and Kolter \cite{fathony2020ap} which has, at best, cubic runtime.
In practice however, optimizing for a metric with $\mathcal{H}$ over all samples in each batch could lead to increased run time due to the number of constant-time operations required to compute the metric.
To mitigate this and further minimize run time, we observe that our proposed $\mathcal{H}$ can be replaced with an reasonably-sized $O(1)$ lookup table by truncating $p$ to several decimal places and precomputing $\mathcal{H}$ for values of $p$ and $\tau$ over the range $[0,1]$.
For example, if an interval between values of $\tau$ is set to $0.1$ and $p$ is truncated at two decimal places, the lookup table has only $1,000$ elements. Using 8-bit storage, this table consumes $1$kB of memory.

\section{Experimental Setup}
\label{supp:sec:experimental-setup}

\subsection{Datasets}
\label{supp:sec:data}

We report results for experiments on four binary classification datasets of tabular data with varying levels of class imbalance applicable to a wide range of domains (Section \ref{sec:direct-optimization-experiment} of the main paper). We also report results for experiments on two different binary classification datasets created from a commonly used image dataset (Section \ref{sec:direct-optimization-experiment-image} of the main paper).

For tabular datasets, features were centered and scaled to unit variance, and the data was split into separate train (64\%), test (20\%) and validation (16\%) sets. For image datasets, the data was normalized and split into separate train (67\%), test (17\%), and validation (17\%) sets.  

\subsubsection{Synthetic Datasets}

Two synthetic datasets were generated, named ``Synthetic 33\%'' ``Synthetic 5\%'' with a positive to negative sample balance of 33\% and 4.76\%, respectively.
Each dataset was formed by creating two isotropic gaussian blobs and removing a randomly-sampled proportion of the positive data points.

\subsubsection{CocktailParty Dataset}

The CocktailParty Dataset\footnote{https://tev.fbk.eu/technologies/cocktailparty-dataset-multi-view-dataset-social-behavior-analysis} \cite{zen2010space} consists of annotations for the locations and orientations of six people in a physical space. The task is to predict whether or not two individuals are part of the same conversational group.
We use a total of 22 spatial properties describing the locations and orientations of the two individuals as well as those of the other four individuals around them.
These features were extracted for every pair of individuals in every frame of the original dataset, resulting in 4800 samples being obtained from 320 frames ($320\times15$ possible pairings = 4800 data points). The spatial coordinates of each data point were chosen such that the individuals in the pair lie along the x-axis with the origin halfway between them. The final dataset has a 30.29\% positive class balance.

\subsubsection{Adult Data Set}

The Adult Data Set\footnote{https://archive.ics.uci.edu/ml/machine-learning-databases/adult/} from the UCI Machine Learning Repository \cite{Dua:2019} consists of data extracted from the 1994 Census database with the intended task of predicting whether a person makes more than \$50K per year. This dataset has 14 features of which 1 feature, indicating the number of people represented by the data point, was removed as it has no bearing on the labeled outcome. This dataset contains 48842 points of which 11687 are positive resulting in a 23.93\% positive class balance.

\subsubsection{Mammography Dataset}

The binary classification data for microcalcifications in the Mammography dataset \cite{woods1993comparative}, available from OpenML,\footnote{https://www.openml.org/d/310} is composed of 6 features, all of which were considered in our experiments. This dataset has 11183 total samples, 260 of which are positive making it imbalanced. The dataset has only 2.32\% positive-class examples.

\subsubsection{Kaggle Credit Card Fraud Detection}

The Kaggle Credit Card Fraud Detection dataset\footnote{https://www.kaggle.com/mlg-ulb/creditcardfraud} consists of transaction data for European card users over two days in September of 2013. It has data for 284807 total transactions and 492 instances of fraud. These positive samples, corresponding to cases of fraud, result in a 0.17\% positive-sample balance. The Kaggle dataset has 28 unnamed features as well as two more named features: transaction time and amount. The time feature was removed to avoid learning correlation between time step and label. Amount was log-scaled due to the wide range in values resulting in a total of 29 features.

\subsubsection{Image Datasets}
The CIFAR-10 dataset\footnote{https://www.cs.toronto.edu/~kriz/cifar.html} \cite{krizhevsky2009learning} contains sixty thousand images evenly distributed across ten mutually exclusive classes: airplane, automobile, bird, cat, deer, dog, frog, horse, ship, and truck. Each image is a three-channel color (RGB) image with a size of 32x32 pixels. We created two binary datasets from CIFAR-10, discussed below.

\textbf{CIFAR-10-Transportation: }
CIFAR-10-Transportation was created to achieve 40\% class balance. If the image was labeled as an airplane, automobile, ship, or truck, it was considered to be in the positive class.

\textbf{CIFAR-10-Frog: }
CIFAR-10-Frog was created to achieve 10\% class balance. Only images labeled as a frog were considered the positive class.

\subsection{Architecture and Training}
\label{supp:sec:arch-train}
Our experiments aim to fairly evaluate our method using different objective functions. Therefore, the same network architecture and training scheme was used unless otherwise noted.
The binary classifier for tabular data was a feedforward neural network consisting of three fully connected layers of 32 units, 16 units, and 1 unit. The first two layers were activated via Rectified Linear Unit (ReLU) \cite{nair2010rectified} and followed by dropout \cite{hinton2012improving}. The final layer was a sigmoid-activated single-unit output.
For image datasets, the Tiny Darknet \cite{redmon2018yolov3} architecture was used.
The ADAM optimizer \cite{kingma2014adam} was used for training with $lr=0.001$ and batch size of 1024 for tabular datasets, and $lr=0.0001$ and batch size of 128 for image datasets. The same batch size and learning rate were used for all methods except \cite{fathony2020ap}, which used hyperparameters (e.g., batch size of 20) suggested by the first author through personal communication. Early stopping was used to terminate training after the validation loss stopped decreasing over a sliding window of 100 epochs.
Each model optimized via a specific loss was trained using PyTorch.
We provide open source implementations of our proposed approach.
Training systems used either an NVIDIA Titan X or RTX 2080ti GPU with an Intel i7 3.7GHz processor and 32GB of RAM, with the exception of the adversarial approach to optimize $F_1$-Score \cite{fathony2020ap}, which had to be run on a CPU due to the author's provided implementation.

\subsection{Hyperparameters}
\label{supp:sec:hyperparameters}

Our method introduces two new hyperparameters: $\tau$, and $k$ or $\delta$.
We empirically determined values of $\tau=0.5$, $k=10$ for $\mathcal{H}^s$, and $\delta=0.1$ for $\mathcal{H}^l$, which were used for all experiments.
Given the results of these experiments on a wide range of problems, we believe these values work well in practice and, therefore, do not require extra effort for tuning them. 

The incorporation of an approximation for the Heaviside step function introduces the $k$ or $\delta$ parameter, depending on which approximation is chosen. For the sigmoid approximation, as $k$ decreases, the derivative of $\mathcal{H}^s$ becomes smoother, facilitating gradient descent. However, in the limit, the approximation may no longer approach $H$ as discussed previously in Section \ref{supp:sec:heaviside-approx-sigmoid} of the Supplementary Material.
Similar to the parameter $k$ in the sigmoid approximation, we define a slope parameter, $\delta$, for our proposed $\mathcal{H}^l$. 
A larger $\delta$ provides a smoother derivative but further deviation from $H$. However, unlike $\mathcal{H}^s$, $\mathcal{H}^l$ always approaches $H$ in the limit. Therefore, the particular choice of $\delta$ is less crucial within a reasonable range. We empirically determined that $0.1 \leq \delta \leq 0.2$ works well in practice.

In order to determine an appropriate batch size, we trained and evaluated models on Accuracy and $F_{1}$-Score using our method over $10$ trials.
Both the sigmoid and the piecewise linear Heaviside approximations were used.
We considered batch sizes in $\{128,1024,2048,4096\}$ and results are shown in Table \ref{supp:tbl:combined-batch}.
Training batch size had a minimal effect on final classifier performance in our experiments and the impact of the choice of approximation varied with class imbalance.
The performance of $\mathcal{H}^l$ and $\mathcal{H}^s$ were similar for the more balanced dataset (Synthetic 33\%).
However, in the case of imbalanced data (Synthetic 5\%), the performance of the $\mathcal{H}^l$ was greater than of $\mathcal{H}^s$ when optimizing $F_1$ over the soft-set confusion matrix and evaluating with the $F_1$-Score metric.
For optimizing Accuracy with our approach, performance evaluated on $F_1$-Score was zero, due to the network's incentive to maximize Accuracy by predicting only dominant-class samples \cite{he2009learning}.
We believe that the advantage of the $\mathcal{H}^l$ in some cases is due to its adherence to the key properties mentioned in Section \ref{sec:heaviside-approximation} of the main paper.
For the experiments on tabular data, we chose a batch size of 1024.

\section{Experiments on Tabular Data with Sigmoid Heaviside Approximation}
\label{supp:sec:direct-optimization-experiment}

Section \ref{sec:direct-optimization-experiment} of the main paper presented results with our method by approximating the Heaviside approximation with our proposed linearly interpolated function.
Table \ref{supp:tbl:direct-results} present results using the sigmoid approximation.

\begin{table*}[tb!]
\setlength{\tabcolsep}{4pt}
\caption{
  Losses (rows): $F_1$, Accuracy ($Acc$), and AUROC ($ROC$) via the proposed method (*) using the sigmoid approximation;
  $F_1$-Score\dag\ via adversarial approach \cite{fathony2020ap} and AUROC\ddag\ via WMW statistic \cite{yan2003optimizing}.
}
\begin{center}
{\small
\begin{tabular}{lrcccccc}
\toprule
\multicolumn{2}{c}{} & \multicolumn{3}{c}{\textbf{CocktailParty} ($\mu\pm\sigma$)} & \multicolumn{3}{c}{\textbf{Adult} ($\mu\pm\sigma$)} \\ \cmidrule(lr){3-5} \cmidrule(lr){6-8}
& Loss & $F_1$-Score & Accuracy & AUROC&$F_1$-Score & Accuracy & AUROC \\ \midrule
(1) & $\text{F}_1*$ & $0.73 \pm 0.02$ & $0.84 \pm 0.01$ & $0.81 \pm 0.01$ & $0.61 \pm 0.04$ & $0.74 \pm 0.07$ & $0.77 \pm 0.03$ \\
(2) & $\text{F}_1$\dag & $0.30 \pm 0.06$ & $0.76 \pm 0.01$ & $0.60 \pm 0.02$ & $0.16 \pm 0.02$ & $0.78 \pm 0.00$ & $0.55 \pm 0.01$ \\
(3) & $\text{Accuracy}*$ & $0.71 \pm 0.02$ & $0.85 \pm 0.01$ & $0.78 \pm 0.01$ & $0.36 \pm 0.03$ & $0.81 \pm 0.00$ & $0.61 \pm 0.01$ \\
(4) & $\text{AUROC}*$ & $0.55 \pm 0.02$ & $0.48 \pm 0.01$ & $0.60 \pm 0.00$ & $0.46 \pm 0.01$ & $0.42 \pm 0.00$ & $0.59 \pm 0.01$ \\
(5) & AUROC\ddag & $0.01 \pm 0.03$ & $0.70 \pm 0.03$ & $0.50 \pm 0.00$ & $0.00 \pm 0.00$ & $0.76 \pm 0.00$ & $0.50 \pm 0.00$ \\
(6) & BCE & $0.70 \pm 0.02$ & $0.85 \pm 0.01$ & $0.78 \pm 0.01$ & $0.26 \pm 0.06$ & $0.80 \pm 0.01$ & $0.58 \pm 0.02$ \\
\end{tabular}
}
{\small
\begin{tabular}{lrcccccc}
\toprule
\multicolumn{2}{c}{} & \multicolumn{3}{c}{\textbf{Mammography} ($\mu\pm\sigma$)} & \multicolumn{3}{c}{\textbf{Kaggle} ($\mu\pm\sigma$)} \\ \cmidrule(lr){3-5} \cmidrule(lr){6-8}
& Loss & $F_1$-Score & Accuracy & AUROC&$F_1$-Score & Accuracy & AUROC \\ \midrule
(1) & $\text{F}_1*$ & $0.50 \pm 0.29$ & $0.89 \pm 0.16$ & $0.75 \pm 0.14$ & $0.81 \pm 0.02$ & $1.00 \pm 0.00$ & $0.89 \pm 0.02$ \\
(2) & $\text{F}_1$\dag & $0.46 \pm 0.08$ & $0.98 \pm 0.00$ & $0.66 \pm 0.04$ & $0.76 \pm 0.06$ & $1.00 \pm 0.00$ & $0.83 \pm 0.04$ \\
(3) & $\text{Accuracy}*$ & $0.00 \pm 0.00$ & $0.98 \pm 0.00$ & $0.50 \pm 0.00$ & $0.72 \pm 0.25$ & $1.00 \pm 0.00$ & $0.84 \pm 0.12$ \\
(4) & $\text{AUROC}*$ & $0.22 \pm 0.01$ & $0.34 \pm 0.00$ & $0.63 \pm 0.01$ & $0.25 \pm 0.01$ & $0.33 \pm 0.00$ & $0.64 \pm 0.00$ \\
(5) & AUROC\ddag & $0.00 \pm 0.01$ & $0.88 \pm 0.12$ & $0.50 \pm 0.00$ & $0.00 \pm 0.00$ & $0.93 \pm 0.15$ & $0.50 \pm 0.00$ \\
(6) & BCE & $0.56 \pm 0.11$ & $0.99 \pm 0.00$ & $0.71 \pm 0.06$ & $0.50 \pm 0.33$ & $1.00 \pm 0.00$ & $0.73 \pm 0.16$ \\
\end{tabular}
}
\end{center}
\label{supp:tbl:direct-results}
\end{table*}

\section{Additional Results}
\label{supp:sec:additional-results}

\subsection{Batch Size}
\label{supp:sec:batch-size}

Table \ref{supp:tbl:combined-batch} provides the results of our experiments exploring the effect of batch size on the performance of our method, using both approximations, compared to the BCE baseline.
This experiment informed our choice of the batch size hyperparameter as described in Section \ref{supp:sec:hyperparameters} of the Supplementary Material.
Apart from zero $F_1$-Score for optimizing Accuracy with soft sets on the Synthetic 5\% dataset, the results are similar within each metric regardless of dataset, batch size, and Heaviside approximation. Performance on the Synthetic \%05 dataset, optimizing $F_1^l$ with soft sets for all batch sizes is comparable, but better than $F_1^s$.
%
\begin{table}[]
    \caption{Accuracy (\textit{Acc}) and $F_1$ (\textit{F}) loss (rows) via the proposed method (*) with the sigmoid ($s$) and linear ($l$) approximations by batch sizes $\{128, 1024, 2048, 4096\}$. The Synthetic 5\% dataset $F_1$-Score is zero when trained with Accuracy  over soft sets loss. Other results are similar within each metric. See text for details.}
    \label{supp:tbl:combined-batch}
    \centering
{\small
\begin{tabular}{lrccccc}
\toprule
\multicolumn{6}{c}{\textbf{Synthetic 5\% Dataset}} \\
\midrule
\multicolumn{2}{c}{} & \multicolumn{4}{c}{Accuracy ($\mu\pm\sigma$)} \\
\cmidrule(lr){3-6}
& \textit{Loss} & $B=128$ & 1024 & 2048 & 4096 \\
\midrule
(1) & $\text{F}^l_1$* & $0.95 \pm 0.00$ & $0.95 \pm 0.01$ & $0.95 \pm 0.01$ & $0.95 \pm 0.01$ \\
(2) & $\text{F}^s_1$* & $0.84 \pm 0.16$ & $0.78 \pm 0.18$ & $0.77 \pm 0.18$ & $0.74 \pm 0.17$ \\
(3) & $\text{Accuracy}^l$* & $0.95 \pm 0.00$ & $0.95 \pm 0.01$ & $0.95 \pm 0.01$ & $0.95 \pm 0.01$ \\
(4) & $\text{Accuracy}^s$* & $0.95 \pm 0.01$ & $0.95 \pm 0.00$ & $0.95 \pm 0.01$ & $0.96 \pm 0.01$ \\
(5) & BCE & $0.96 \pm 0.01$ & $0.95 \pm 0.01$ & $0.95 \pm 0.01$ & $0.95 \pm 0.00$ \\
\midrule
\multicolumn{2}{c}{} & \multicolumn{4}{c}{$F_1$-Score ($\mu\pm\sigma$)} \\
\cmidrule(lr){3-6}
& \textit{Loss} & $B=128$ & 1024 & 2048 & 4096 \\
\midrule
(1) & $\text{F}^l_1$* & $0.52 \pm 0.04$ & $0.50 \pm 0.06$ & $0.52 \pm 0.05$ & $0.48 \pm 0.04$ \\
(2) & $\text{F}^s_1$* & $0.32 \pm 0.18$ & $0.27 \pm 0.22$ & $0.24 \pm 0.20$ & $0.21 \pm 0.17$ \\
(3) & $\text{Accuracy}^l$* & $0.00 \pm 0.00$ & $0.00 \pm 0.00$ & $0.00 \pm 0.00$ & $0.00 \pm 0.00$ \\
(4) & $\text{Accuracy}^s$* & $0.00 \pm 0.00$ & $0.00 \pm 0.00$ & $0.00 \pm 0.00$ & $0.00 \pm 0.00$ \\
(5) & BCE & $0.22 \pm 0.10$ & $0.21 \pm 0.11$ & $0.12 \pm 0.08$ & $0.09 \pm 0.06$ \\
\end{tabular}
}
{\small
\begin{tabular}{lrccccc}
\toprule
\multicolumn{6}{c}{\textbf{Synthetic 33\% Dataset}} \\
\midrule
\multicolumn{2}{c}{} & \multicolumn{4}{c}{Accuracy ($\mu\pm\sigma$)} \\
\cmidrule(lr){3-6}
& \textit{Loss} & $B=128$ & 1024 & 2048 & 4096 \\
\midrule
(1) & $\text{F}^l_1$* & $0.84 \pm 0.01$ & $0.84 \pm 0.01$ & $0.84 \pm 0.00$ & $0.85 \pm 0.01$ \\
(2) & $\text{F}^s_1$* & $0.84 \pm 0.01$ & $0.83 \pm 0.01$ & $0.83 \pm 0.02$ & $0.83 \pm 0.01$ \\
(3) & $\text{Accuracy}^l$* & $0.85 \pm 0.01$ & $0.85 \pm 0.01$ & $0.86 \pm 0.01$ & $0.85 \pm 0.01$ \\
(4) & $\text{Accuracy}^s$* & $0.85 \pm 0.00$ & $0.85 \pm 0.01$ & $0.85 \pm 0.01$ & $0.85 \pm 0.01$ \\
(5) & BCE & $0.84 \pm 0.01$ & $0.85 \pm 0.01$ & $0.84 \pm 0.01$ & $0.85 \pm 0.01$ \\
\midrule
\multicolumn{2}{c}{} & \multicolumn{4}{c}{$F_1$-Score ($\mu\pm\sigma$)} \\
\cmidrule(lr){3-6}
& \textit{Loss} & $B=128$ & 1024 & 2048 & 4096 \\
\midrule
(1) & $\text{F}^l_1$* & $0.77 \pm 0.01$ & $0.78 \pm 0.01$ & $0.78 \pm 0.01$ & $0.78 \pm 0.01$ \\
(2) & $\text{F}^s_1$* & $0.77 \pm 0.01$ & $0.76 \pm 0.01$ & $0.77 \pm 0.02$ & $0.77 \pm 0.01$ \\
(3) & $\text{Accuracy}^l$* & $0.77 \pm 0.01$ & $0.77 \pm 0.01$ & $0.78 \pm 0.01$ & $0.78 \pm 0.01$ \\
(4) & $\text{Accuracy}^s$* & $0.77 \pm 0.01$ & $0.77 \pm 0.01$ & $0.77 \pm 0.01$ & $0.77 \pm 0.01$ \\
(5) & BCE & $0.74 \pm 0.01$ & $0.74 \pm 0.02$ & $0.74 \pm 0.01$ & $0.74 \pm 0.02$ \\
\end{tabular}
}
\end{table}

\subsection{Balancing between Precision and Recall}
\label{supp:sec:precision-recall}

Experimental results for balancing between precision and recall as described in Section \ref{sec:precision-recall} of the main paper are reported in Table \ref{supp:tbl:precision-recall} for the remaining three tabular datasets.
These supplementary results are similar to those presented in the main paper.

\begin{table}[tb!]
  \caption{Optimizing $F_\beta$-Scores ($\beta = \{1,2,3\}$) via our method to balance between precision and recall while maximizing $F_1$-Score.}
  \label{supp:tbl:precision-recall}
    \centering
{\small
\begin{tabular}{lrccccc}
\toprule
\multicolumn{7}{c}{Adult} \\ \midrule
& Loss & $F_1$-Score & $F_2$-Score & $F_3$-Score & Precision & Recall \\ \midrule
(1) & $\text{F}_1$* & $0.64 \pm 0.01$ & $0.72 \pm 0.03$ & $0.75 \pm 0.05$ & $0.54 \pm 0.04$ & $0.78 \pm 0.07$ \\
(2) & $\text{F}_2$* & $0.43 \pm 0.02$ & $0.65 \pm 0.02$ & $0.79 \pm 0.01$ & $0.27 \pm 0.02$ & $1.00 \pm 0.00$ \\
(3) & $\text{F}_3$* & $0.39 \pm 0.01$ & $0.61 \pm 0.01$ & $0.76 \pm 0.01$ & $0.24 \pm 0.01$ & $1.00 \pm 0.00$ \\
\end{tabular}
}
{\small
\begin{tabular}{lrccccc}
\toprule
\multicolumn{7}{c}{Kaggle} \\ \midrule
& Loss & $F_1$-Score & $F_2$-Score & $F_3$-Score & Precision & Recall \\ \midrule
(1) & $\text{F}_1*$ & $0.82 \pm 0.04$ & $0.80 \pm 0.04$ & $0.80 \pm 0.04$ & $0.84 \pm 0.04$ & $0.79 \pm 0.04$ \\
(2) & $\text{F}_2*$ & $0.81 \pm 0.02$ & $0.80 \pm 0.03$ & $0.80 \pm 0.03$ & $0.83 \pm 0.02$ & $0.79 \pm 0.03$ \\
(3) & $\text{F}_3*$ & $0.82 \pm 0.03$ & $0.81 \pm 0.04$ & $0.81 \pm 0.04$ & $0.82 \pm 0.04$ & $0.81 \pm 0.04$ \\
\end{tabular}
}
{\small
\begin{tabular}{lrccccc}
\toprule
\multicolumn{7}{c}{CocktailParty} \\ \midrule
& Loss & $F_1$-Score & $F_2$-Score & $F_3$-Score & Precision & Recall \\ \midrule
(1) & $\text{F}_1*$ & $0.74 \pm 0.01$ & $0.75 \pm 0.02$ & $0.75 \pm 0.02$ & $0.73 \pm 0.03$ & $0.75 \pm 0.02$ \\
(2) & $\text{F}_2*$ & $0.68 \pm 0.02$ & $0.80 \pm 0.01$ & $0.85 \pm 0.01$ & $0.54 \pm 0.03$ & $0.91 \pm 0.01$ \\
(3) & $\text{F}_3*$ & $0.60 \pm 0.02$ & $0.77 \pm 0.01$ & $0.85 \pm 0.01$ & $0.43 \pm 0.02$ & $0.95 \pm 0.02$ \\
\end{tabular}
}

\end{table}

\section{Additional Comparisons with Other Methods of Binary Classification}
\label{supp:sec:additional-experiments}

\subsection{Other Baselines}
\label{supp:ssec:experiments:other}
We compare our method with other baselines on the tabular datasets, shown in Table \ref{tbl:addl-direct-results}.
Included in these supplementary results are neural network classifiers trained on Dice \cite{milletari2016v}, 
and SVM\textsuperscript{perf} \cite{joachims2006training} to which \cite{narasimhan2014statistical} compares.
SVM\textsuperscript{perf} is trained on Errorrate loss (E), F$_1$ loss (F$_1$), and AUROC loss (ROC).
Our method outperforms the Dice loss in all cases.
Our method is comparable or better than SVM\textsuperscript{perf}.

\begin{table*}[bt!]
\setlength{\tabcolsep}{4pt}
\caption{Other baselines on tabular data. See text for details.
}
\begin{center}
\small
\begin{tabular}{lccccccc}
\toprule
\multicolumn{2}{c}{} & \multicolumn{3}{c}{\textbf{CocktailParty} ($\mu\pm\sigma$)} & \multicolumn{3}{c}{\textbf{Adult} ($\mu\pm\sigma$)} \\ \cmidrule(lr){3-5} \cmidrule(lr){6-8}
& \textit{Loss} & Accuracy & $F_1$-Score & AUROC & Accuracy & $F_1$-Score & AUROC \\ \midrule
(1) & Dice & $0.68 \pm 0.04$ & $0.49 \pm 0.02$ & $0.56 \pm 0.02$ & $0.50 \pm 0.15$ & $0.40 \pm 0.02$ & $0.55 \pm 0.02$ \\
(2) & $SVM_{E}^{perf}$ & $0.82$ & $0.66$ & $0.75$ & $0.81$ & $0.60$ & $0.75$ \\
(3) & $SVM_{F1}^{perf}$ & $0.78$ & $0.69$ & $0.78$ & $0.51$ & $0.48$ & $0.66$ \\
(4) & $SVM_{ROC}^{perf}$ & $0.76$ & $0.67$ & $0.77$ & $0.32$ & $0.41$ & $0.56$ \\
\bottomrule
\end{tabular}
\end{center}
\begin{center}
{\small
\begin{tabular}{lccccccc}
\multicolumn{2}{c}{} & \multicolumn{3}{c}{\textbf{Mammography} ($\mu\pm\sigma$)} & \multicolumn{3}{c}{\textbf{Kaggle} ($\mu\pm\sigma$)} \\ \cmidrule(lr){3-5} \cmidrule(lr){6-8}
& \textit{Loss} & Accuracy & $F_1$-Score & AUROC & Accuracy & $F_1$-Score & AUROC \\ \midrule
(1) & Dice & $0.85 \pm 0.17$ & $0.19 \pm 0.09$ & $0.65 \pm 0.06$ & $0.67 \pm 0.15$ & $0.10 \pm 0.06$ & $0.76 \pm 0.06$ \\
(2) & $SVM_{E}^{perf}$ & $0.98$ & $0.51$ & $0.68$ & $1.00$ & $0.80$ & $0.89$ \\
(3) & $SVM_{F1}^{perf}$ & $0.98$ & $0.59$ & $0.82$ & $1.00$ & $0.80$ & $0.90$ \\
(4) & $SVM_{ROC}^{perf}$ & $0.61$ & $0.11$ & $0.77$ & $0.67$ & $0.01$ & $0.83$ \\
\bottomrule
\end{tabular}
}
\end{center}
\label{tbl:addl-direct-results}
\end{table*}

\subsection{Weighted Loss Results}

A common method of dealing with sample imbalance is weighting. In this section, we show that our method can also be used with weighting. During training, weighted the loss by the amount of class imbalance in each dataset and compared our method with binary cross entropy. More specifically, we computed sample weights $W$ for each dataset between negative ($n$) and positive ($p$) samples where positive samples always correspond to the minority class:
\begin{equation*}
  \begin{aligned}
    W_n = \frac{1}{|n|}\frac{|n|+|p|}{2.0} \quad \quad
    W_p = \frac{1}{|p|}\frac{|n|+|p|}{2.0}
  \end{aligned}
\end{equation*}
The weights calculated for our datasets are shown in Table \ref{tlb:cls-weight}.

\begin{table*}[bt!]
\begin{center}
\caption{Dataset sample weights. See text for details.}
\begin{tabular}{rcc}
\toprule
Dataset & Negative & Positive \\
\midrule
CocktailParty & 0.72 & 1.65 \\
Adult & 0.66 & 2.07 \\
Mammography & 0.51 & 21.55 \\
Kaggle & 0.50 & 290.25 \\
\bottomrule
\end{tabular}
\label{tlb:cls-weight}
\end{center}
\end{table*}

Table \ref{tbl:weighted-results} shows losses (rows): $F_1$ and Accuracy ($Acc$) trained with the linear ($l$) and sigmoid ($s$) approximations compared with the traditional binary cross-entropy.
Compared to Table 1 in the main paper, in Table \ref{tbl:weighted-results} the $F_1$ loss (line 1) decreased in performance and Accuracy loss (line 2) increased in performance.
In all cases, our method without weighting performs similarly or better than the BCE baseline with weighting. 

Another method of dealing with class imbalance is oversampling. Oversampling is when samples are repeatedly sampled from the minority class until class balance is reached. We applied oversampling to the training split of each dataset. Using this technique we achieved a positive versus negative sample split in each dataset nearer 50/50, detailed in Table \ref{tbl:oversampling}.

The performance of our method versus the BCE baseline, both with oversampling, is shown in Table \ref{tbl:oversampling-results}. In general our method performs better without oversampling.
The BCE baseline shows improvement in some cases compared to BCE with weighting.
In all cases, our method without oversampling performs similarly or better than the BCE baseline with oversampling. 

\begin{table*}[bt!]
\begin{center}
\setlength{\tabcolsep}{4pt}
\caption{Weighted loss results. See text for details.}
{\small
\begin{tabular}{lrcccc}
\toprule
\multicolumn{2}{c}{} & \multicolumn{2}{c}{\textbf{CocktailParty} ($\mu\pm\sigma$)} & \multicolumn{2}{c}{\textbf{Adult} ($\mu\pm\sigma$)} \\ \cmidrule(lr){3-4} \cmidrule(lr){5-6}
& Loss & $F_1$-Score & Accuracy&$F_1$-Score & Accuracy \\ \midrule
(1) & $\text{F}^l_1$ & $0.72 \pm 0.01$ & $0.80 \pm 0.01$ & $0.45 \pm 0.03$ & $0.41 \pm 0.06$ \\
(2) & $\text{F}^s_1$ & $0.71 \pm 0.03$ & $0.77 \pm 0.06$ & $0.42 \pm 0.02$ & $0.36 \pm 0.05$ \\
(3) & $\text{Accuracy}^l$ & $0.75 \pm 0.02$ & $0.83 \pm 0.01$ & $0.57 \pm 0.04$ & $0.66 \pm 0.06$ \\
(4) & $\text{Accuracy}^s$ & $0.72 \pm 0.02$ & $0.82 \pm 0.02$ & $0.54 \pm 0.03$ & $0.61 \pm 0.06$ \\
(5) & BCE & $0.75 \pm 0.02$ & $0.85 \pm 0.01$ & $0.56 \pm 0.05$ & $0.80 \pm 0.02$ \\
\end{tabular}
}
{\small
\begin{tabular}{lrcccc}
\toprule
\multicolumn{2}{c}{} & \multicolumn{2}{c}{\textbf{Mammography} ($\mu\pm\sigma$)} & \multicolumn{2}{c}{\textbf{Kaggle} ($\mu\pm\sigma$)} \\ \cmidrule(lr){3-4} \cmidrule(lr){5-6}
& Loss & $F_1$-Score & Accuracy&$F_1$-Score & Accuracy \\ \midrule
(1) & $\text{F}^l_1$ & $0.31 \pm 0.04$ & $0.90 \pm 0.01$ & $0.13 \pm 0.02$ & $0.98 \pm 0.01$ \\
(2) & $\text{F}^s_1$ & $0.28 \pm 0.05$ & $0.88 \pm 0.04$ & $0.41 \pm 0.17$ & $0.99 \pm 0.01$ \\
(3) & $\text{Accuracy}^l$ & $0.34 \pm 0.04$ & $0.92 \pm 0.01$ & $0.42 \pm 0.08$ & $1.00 \pm 0.00$ \\
(4) & $\text{Accuracy}^s$ & $0.35 \pm 0.04$ & $0.92 \pm 0.01$ & $0.38 \pm 0.11$ & $0.99 \pm 0.00$ \\
(5) & BCE & $0.43 \pm 0.06$ & $0.95 \pm 0.01$ & $0.21 \pm 0.05$ & $0.99 \pm 0.00$ \\
\end{tabular}
}

\label{tbl:weighted-results}
\end{center}
\end{table*}

\begin{table*}[bt!]
\begin{center}
\caption{Data balancing via Oversampling. See text for details.}
\begin{tabular}{rcccccc}
\toprule
& \multicolumn{3}{c}{Entire Dataset} & \multicolumn{3}{c}{Oversampled Train Split} \\ \cmidrule(lr){2-4}  \cmidrule(lr){5-7}
Dataset & Total & \multicolumn{2}{c}{Positive} & Total & \multicolumn{2}{c}{Positive} \\
\midrule
CocktailParty & 4800 & 1454 & 30.29\% & 4284 & 2142 & 50\% \\
Adult & 48842 & 11687 & 23.93\% & 39564 & 19782 & 50\% \\
Mammography & 11183 & 260 & 2.32\% & 13970 & 6985 & 50\%  \\
Kaggle & 284807 & 492 & 0.17\% & 363894 & 181947 & 50\% \\
\bottomrule
\end{tabular}
\label{tbl:oversampling}
\end{center}
\end{table*}

\begin{table*}[bt!]
\begin{center}
\setlength{\tabcolsep}{4pt}
\caption{Oversampling Results. See text for details.}

{\small
\begin{tabular}{lrcccc}
\toprule
\multicolumn{2}{c}{} & \multicolumn{2}{c}{\textbf{CocktailParty} ($\mu\pm\sigma$)} & \multicolumn{2}{c}{\textbf{Adult} ($\mu\pm\sigma$)} \\ \cmidrule(lr){3-4} \cmidrule(lr){5-6}
& Loss & $F_1$-Score & Accuracy&$F_1$-Score & Accuracy \\ \midrule
(1) & $\text{F}^l_1$ & $0.72 \pm 0.02$ & $0.80 \pm 0.01$ & $0.45 \pm 0.02$ & $0.42 \pm 0.05$ \\
(2) & $\text{F}^s_1$ & $0.71 \pm 0.02$ & $0.79 \pm 0.01$ & $0.43 \pm 0.02$ & $0.39 \pm 0.04$ \\
(3) & $\text{Accuracy}^l$ & $0.74 \pm 0.02$ & $0.83 \pm 0.02$ & $0.55 \pm 0.06$ & $0.62 \pm 0.10$ \\
(4) & $\text{Accuracy}^s$ & $0.73 \pm 0.03$ & $0.82 \pm 0.02$ & $0.53 \pm 0.04$ & $0.59 \pm 0.07$ \\
(5) & BCE & $0.75 \pm 0.02$ & $0.85 \pm 0.01$ & $0.59 \pm 0.03$ & $0.81 \pm 0.02$ \\
\end{tabular}
}
{\small
\begin{tabular}{lrcccc}
\toprule
\multicolumn{2}{c}{} & \multicolumn{2}{c}{\textbf{Mammography} ($\mu\pm\sigma$)} & \multicolumn{2}{c}{\textbf{Kaggle} ($\mu\pm\sigma$)} \\ \cmidrule(lr){3-4} \cmidrule(lr){5-6}
& Loss & $F_1$-Score & Accuracy&$F_1$-Score & Accuracy \\ \midrule
(1) & $\text{F}^l_1$ & $0.38 \pm 0.07$ & $0.93 \pm 0.02$ & $0.73 \pm 0.06$ & $1.00 \pm 0.00$ \\
(2) & $\text{F}^s_1$ & $0.06 \pm 0.03$ & $0.48 \pm 0.04$ & $0.69 \pm 0.07$ & $1.00 \pm 0.00$ \\
(3) & $\text{Accuracy}^l$ & $0.49 \pm 0.07$ & $0.96 \pm 0.01$ & $0.75 \pm 0.04$ & $1.00 \pm 0.00$ \\
(4) & $\text{Accuracy}^s$ & $0.49 \pm 0.06$ & $0.96 \pm 0.01$ & $0.74 \pm 0.04$ & $1.00 \pm 0.00$ \\
(5) & BCE & $0.53 \pm 0.04$ & $0.97 \pm 0.00$ & $0.57 \pm 0.06$ & $1.00 \pm 0.00$ \\
\end{tabular}
}

\label{tbl:oversampling-results}
\end{center}
\end{table*}

\subsection{Image Data}
\label{supp:ssec:experiments:image}
We also compared the performance of our proposed method to neural network classifiers trained with Dice as the loss function \cite{milletari2016v}. Results from Dice (logit only) and Dice$^s$ (sigmoid output) are in Table \ref{tbl:addl-results-cifar}. Our method greatly outperformed Dice and DICE$^s$, likely due to the fact that Dice is designed for image segmentation, rather than binary classification. 

\begin{table}
  \caption{Losses (rows): DICE and DICE$^s$ optimize the DICE coefficient. DICE is logit-only, and DICE$^s$ uses a sigmoid output layer.}
  \label{tbl:addl-results-cifar}
  \centering
  {\small
  \begin{tabular}{lrcc}
    \toprule
    \multicolumn{4}{c}{\textbf{CIFAR-10-Transportation} Results ($\mu\pm\sigma$)} \\
    \midrule
    & \textit{Loss} & Accuracy & $F_1$-Score \\
    \midrule
    (6) &$\textit{DICE}$& $ 0.521  \pm 0.02 $ & $ 0.625  \pm 0.01 $ \\
    (7) & $\textit{DICE}^s$  & $ 0.546  \pm 0.04 $ & $ 0.638  \pm 0.02 $ \\
    \arrayrulecolor{black}\bottomrule
  \end{tabular}
  \quad
    \begin{tabular}{lrcc}
  
    \toprule
    \multicolumn{4}{c}{\textbf{CIFAR-10-Frog} Results ($\mu\pm\sigma$)} \\
    \midrule
    & \textit{Loss} & Accuracy & $F_1$-Score \\
    \midrule
    (6) & $\textit{DICE}$  & $ 0.230  \pm 0.02 $ & $ 0.205  \pm 0.00 $ \\
    (7) & $\textit{DICE}^s$  & $ 0.277  \pm 0.04 $ & $ 0.216  \pm 0.01 $ \\
    \arrayrulecolor{black}\bottomrule
  \end{tabular}
  }
\end{table}

\clearpage

\bibliographystyle{plain}
\bibliography{references}

\end{document}